\documentclass{article}

\usepackage[final]{corl_2026} 
\usepackage{amssymb} 
\usepackage{amsmath}
\usepackage{enumitem}
\usepackage{booktabs}
\usepackage{multirow}
\usepackage{multicol}
\usepackage[table,xcdraw]{xcolor}
\usepackage{graphicx}
\usepackage{caption}
\usepackage{wrapfig}
\usepackage{subcaption}

\title{
    \resizebox{0.98\textwidth}{!}{%
        \begin{tabular}{@{}c@{}}
          TDC: Sim-to-Real Transferable Directional Compliance \\[0.2ex]
          for Contact-Rich Manipulation
        \end{tabular}%
    }%
}

\author{
  \textbf{Yifei Yang}\textsuperscript{\normalfont 1}
  \hspace{1.2em}
  \textbf{Anzhe Chen}\textsuperscript{\normalfont 1}
  \hspace{1.2em}
  \textbf{Zhenjie Zhu}\textsuperscript{\normalfont 1}
  \hspace{1.2em}
  \textbf{Kechun Xu}\textsuperscript{\normalfont 1}
  \hspace{1.2em}
  \textbf{Yunxuan Mao}\textsuperscript{\normalfont 1}
  \\[0.3ex]
  \textbf{Yufei Wei}\textsuperscript{\normalfont 1}
  \hspace{1.2em}
  \textbf{Lu Chen}\textsuperscript{\normalfont 2}
  \hspace{1.2em}
  \textbf{Rong Xiong}\textsuperscript{\normalfont 1,2}
  \hspace{1.2em}
  \textbf{Yue Wang}\textsuperscript{\normalfont 1,*}
  \\[0.6ex]
  {\normalfont
    \textsuperscript{1}Zhejiang University}
  \\[0.2ex]
  {\normalfont
    \textsuperscript{2}Zhejiang Humanoid Robot Innovation Center}
  \\[0.2ex]
  {\normalfont
    \textsuperscript{*}Corresponding authors.}
  \\[0.8ex]
  {\normalfont\small
    Project Page:
    \url{https://yifei-y.github.io/project-pages/TDC/}}
}

\begin{document}
\maketitle


\vspace{-0.3cm}
\begin{abstract}
    Contact-rich manipulation requires robots to regulate both motion and interaction forces, yet achieving adaptive compliance remains a fundamental challenge. Learning from real-world data is costly and risky, while simulation-based approaches struggle with the sim-to-real gap in contact dynamics; existing sim-to-real methods either require real-world adaptation or sacrifice adaptive compliance by relying on isotropic compliant controllers. Our key insight is that force regulation decomposes into a time-varying but simulation-transferable directional component and a dynamics-sensitive but manually tunable magnitude component. We instantiate this directional component as two policy outputs, a task frame and a control mode vector, predicted by a visuomotor policy adapted from a pre-trained VLA model and trained via imitation learning on automatically generated simulation demonstrations. At deployment, an admittance controller integrates these predictions with human-specified stiffness and target wrench values to realize adaptive compliance. Our approach achieves adaptive compliance using only simulation data and can benefit from large-scale VLA pre-training. Extensive real-world experiments on four contact-rich tasks, microwave opening, peg-in-hole insertion, whiteboard wiping, and door opening, demonstrate strong task success rates and robustness to external disturbances.
\end{abstract}

\keywords{Contact-Rich Manipulation, Sim-to-Real, Imitation Learning} 


\begin{figure}[h]
    \centering
    \includegraphics[width=\textwidth]{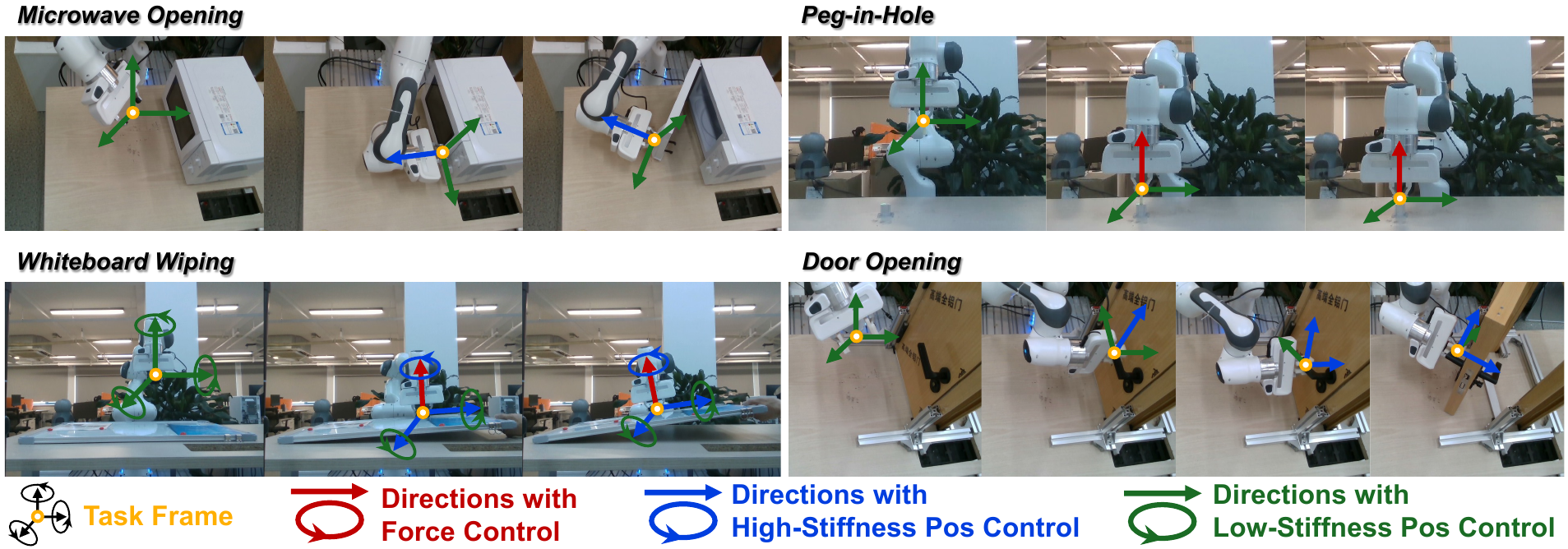}
    \caption{Task frames and per-axis control modes (the directional component of force regulation) visualized across four contact-rich tasks. Rotational axes are omitted for three tasks where they remain in low-stiffness position control. This directional component evolves with task geometry, making manual specification impractical, yet it is governed by geometry rather than contact dynamics and thus sim-to-real transferable. TDC trains a policy on simulation data alone to predict it, achieving adaptive compliance.}
    \label{fig:teaser}
\end{figure}

\section{Introduction} \label{sec:intro}

    Contact-rich manipulation arises in a wide range of robotic applications, from factory assembly~\cite{assembly_1, assembly_2, assembly_3, assembly_4} to everyday household tasks~\cite{household_1, household_2, household_3}. These tasks require robots to not only generate feasible motion trajectories but also regulate interaction forces appropriately. Prior work has achieved remarkable performance by learning such skills from real-world data~\cite{hilserl, acp, rdp, forcemimic, forcevla}; however, real-world data collection for contact-rich tasks is nontrivial, expensive, and carries safety risks. This motivates learning in simulation, yet the sim-to-real gap in contact dynamics severely degrades force regulation upon deployment. Some studies attempt to bridge this gap through real-world adaptation~\cite{adaptation_1, adaptation_2, adaptation_3} or system identification~\cite{si_1, si_2, si_3}, but both still require real-world interaction data. Others~\cite{watch, forge, bridging} adopt reinforcement learning with dynamics randomization, but training visuomotor policies under this paradigm is intractable due to the compounding difficulty of visual and dynamic variation, forcing reliance on state-based policies and precluding the use of robotic foundation models~\cite{e2vla, pi0, pi05}. Finally, some works~\cite{disambiguate, unidoormanip, automate} sidestep the dynamics gap by training position-only policies paired with manually tuned, isotropic compliant controllers, sacrificing adaptive compliance. This raises a central question: \textit{Can we achieve adaptive compliance in contact-rich manipulation with a visuomotor policy using only simulation data?}

    Our key insight is that force regulation can be decomposed into two components: \textit{direction} and \textit{magnitude}. The directional component, i.e., whether each degree of freedom (DoF) should be regulated using force control, high-stiffness position control, or compliant position control, is governed by task geometry rather than contact dynamics, making it sim-to-real transferable, yet it is time-varying and practically impossible to specify manually. The magnitude component, comprising stiffness values and target force magnitudes, is coupled to contact dynamics but remains constant throughout execution and is therefore manually tunable.

    Building on this insight, we propose \textbf{TDC} (\textbf{T}ransferable \textbf{D}irectional \textbf{C}ompliance), a novel sim-to-real framework for contact-rich manipulation. We instantiate the directional component as two policy outputs: a task frame whose axes align with the task's motion constraints, and a control mode vector that specifies the control mode along each task frame axis. Specifically, a visuomotor policy adapted from a pre-trained VLA model is trained with imitation learning on automatically generated simulation data to predict these outputs alongside end-effector pose and gripper command. At deployment, an admittance controller integrates the policy's predictions with human-specified stiffness values and target force magnitudes to realize adaptive compliance. This approach uses only simulation data and can benefit from VLA pre-training. Besides, manual tuning during deployment is minor: a single stiffness value suffices across all our tasks, and Sec.~\ref{sec:results} demonstrates robustness to exact parameter values.

    Building on this insight, we propose \textbf{TDC} (\textbf{T}ransferable \textbf{D}irectional \textbf{C}ompliance), a novel sim-to-real framework for contact-rich manipulation. We instantiate the directional component as two policy outputs: a task frame whose axes align with the task's motion constraints, and a control mode vector that specifies the control mode along each task frame axis. Specifically, a visuomotor policy adapted from a pre-trained VLA model is trained with imitation learning on automatically generated simulation data to predict these outputs alongside end-effector pose and gripper command. At deployment, an admittance controller integrates the policy's predictions with human-specified stiffness values and target force magnitudes to realize adaptive compliance. This approach uses only simulation data and can benefit from VLA pre-training. Besides, manual tuning during deployment is minor: a single stiffness value suffices across all our tasks, and Sec.~\ref{sec:results} demonstrates broad usable ranges of stiffness and target force values.

    We validate TDC on four contact-rich tasks (microwave opening, peg-in-hole insertion, whiteboard wiping, and door opening) spanning diverse interaction constraints. TDC demonstrates strong success rates and robustness to external disturbances. Our main contributions are:
    \begin{itemize}[itemsep=0pt, topsep=0pt]
        \item \textbf{A decomposition of force regulation} into a sim-to-real transferable directional component and a manually tunable magnitude component, enabling adaptive compliance in contact-rich manipulation with only simulation data.
        \item \textbf{A sim-to-real framework} that trains a VLA-based visuomotor policy on automatically generated simulation demonstrations to predict the directional component, paired with an admittance controller that integrates human-specified magnitude parameters.
        \item \textbf{Experimental validation} on four diverse contact-rich tasks, demonstrating strong success rates and robustness to parameter variation and external disturbances.
    \end{itemize}


\section{Related Work}

    \subsection{Real-World Learning for Contact-Rich Manipulation}

        Learning contact-rich manipulation in the real world provides access to accurate contact dynamics without explicit modeling. One line of work adopts imitation learning, either explicitly predicting desired forces~\cite{forcemimic, forcevla2} or stiffness profiles~\cite{acp}, or implicitly representing contact forces as pose offsets executed by compliant controllers~\cite{rdp, forcevla, tavla, foar, factr}. CR-DAgger~\cite{crdagger} trains a residual policy for force-aware corrections via human-in-the-loop interaction. Another line employs real-world reinforcement learning~\cite{hilserl, rl100}, which can discover sophisticated strategies beyond human demonstrations. Despite their effectiveness, these approaches require extensive real-world interaction to acquire force-related supervision, which is nontrivial, costly, and poses safety risks.
 
        Concurrent to our work, Fang et al.~\cite{forcepolicy} propose a similar formulation that predicts a task frame and per-axis control modes paired with a hybrid force-position controller, and introduce a method for extracting task frames from real-world demonstrations. Our approach differs in three respects: we distinguish a third low-stiffness compliant mode to accommodate prediction noise and disturbances; we train entirely on simulation data; and since task frames are geometrically defined, ground-truth labels are directly obtainable in simulation, avoiding extraction noise from real demonstrations.

    \subsection{Sim-to-Real Transfer for Contact-Rich Manipulation}

        To reduce the cost and safety risks of real-world data collection, a large body of work explores sim-to-real transfer for contact-rich manipulation. The central challenge is the sim-to-real mismatch in contact dynamics, which makes force-related behaviors particularly difficult to transfer.

        One line of work addresses this gap through real-world adaptation via residual learning or policy finetuning~\cite{adaptation_1, adaptation_2, adaptation_3, adaptation_4}, or system identification~\cite{si_1, si_2, si_3, si_4}. These approaches facilitate transfer but still require real-world data, and system identification is further limited by simplified contact models that cannot fully capture real contact phenomena.

        A complementary paradigm leverages reinforcement learning with dynamics randomization for zero-shot transfer~\cite{watch, doorman, forge, bridging}. However, applying this to visuomotor policies is intractable due to the compounding difficulty of visual and dynamic variation, and most works therefore rely on state-based policies, forgoing the benefits of robotic foundation models.

        A third class of methods pairs simulation-trained position-only policies with compliant controllers at deployment~\cite{disambiguate, unidoormanip, automate}, enabling safe interaction without accurate contact simulation, but the resulting compliance is typically isotropic and fixed.

        In contrast, our decomposition separates force regulation into a transferable directional component predicted by a visuomotor policy and a manually tunable magnitude component, achieving adaptive compliance without real-world demonstrations while leveraging pre-trained VLA models.


\section{Method}

    \begin{figure}
        \centering
        \includegraphics[width=\textwidth]{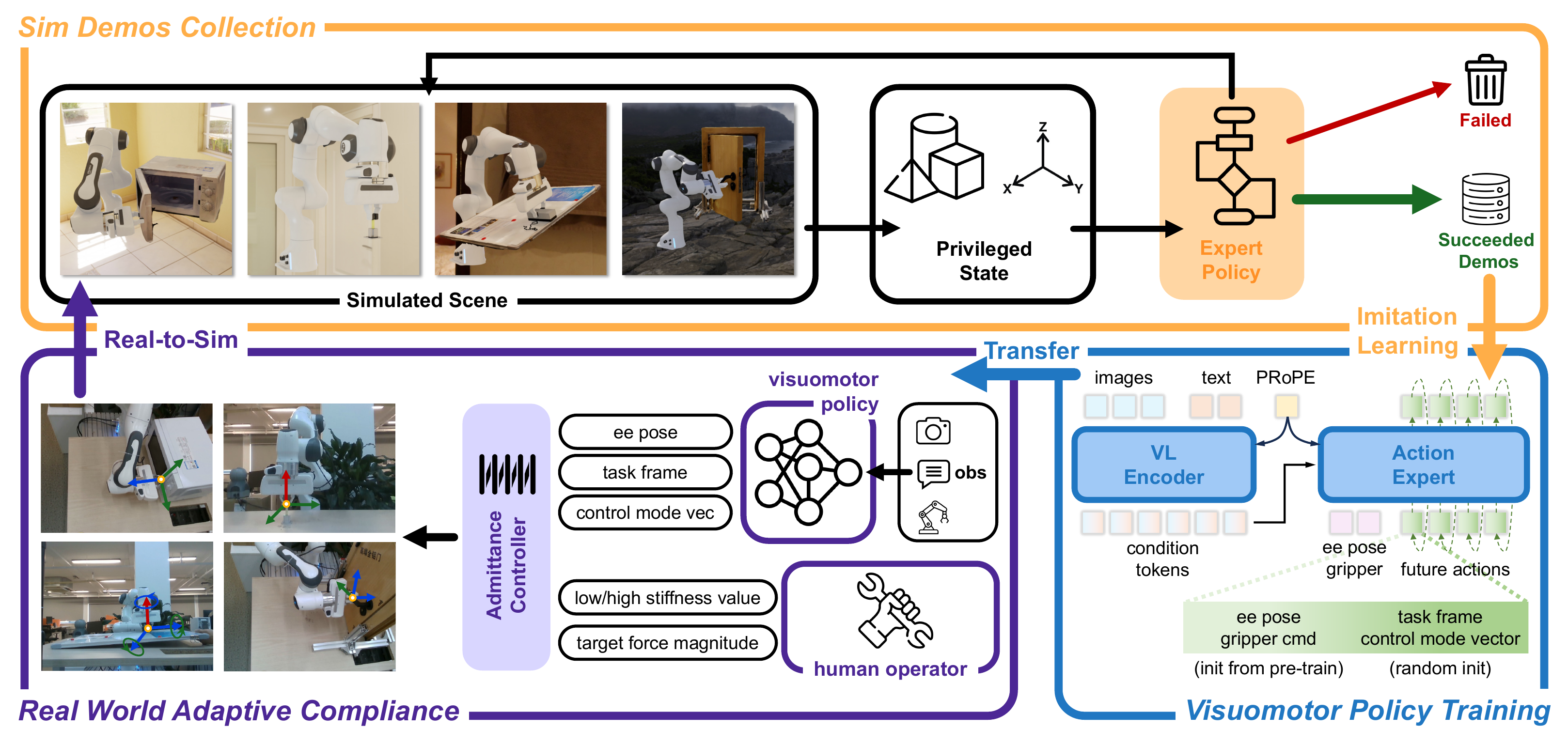}
        \caption{Overview of the proposed framework. (Top) Real objects are scanned into simulated scenes, where a privileged expert generates demonstrations and retains only successful rollouts. (Bottom right) A visuomotor policy built on a pretrained VLA model jointly predicts end-effector poses, task frame orientations, and control mode vectors; new output heads are randomly initialized while original weights are inherited. (Bottom left) At deployment, an admittance controller combines policy predictions with operator-tuned stiffness and force setpoints for adaptive compliance.}
        \label{fig:framework}
    \end{figure}

    Our goal is to achieve adaptive compliance in real-world contact-rich manipulation using only simulation data. Our framework decomposes adaptive compliance into a time-varying but sim-to-real transferable directional component and a dynamics-sensitive but manually tunable magnitude component. We instantiate the directional component as the task frame and control mode vector (Sec.~\ref{sec:taskframe}). A privileged expert policy generates demonstrations in simulation (Sec.~\ref{sec:demo}), and a visuomotor policy adapted from a pre-trained VLA model is trained via imitation learning on these data to predict the task frame and control mode vector (Sec.~\ref{sec:policy}). At deployment, an admittance controller integrates policy predictions with human-specified stiffness and target force values to realize adaptive compliance (Sec.~\ref{sec:controller}). Fig.~\ref{fig:framework} illustrates the overall framework.

    \subsection{Task Frame} \label{sec:taskframe}

        The Task Frame Formalism (TFF)~\cite{taskframe} defines an orthogonal reference frame whose six axes (three translational, three rotational) are partitioned by contact geometry: unconstrained directions are position-controlled while constrained directions are force-controlled. Since the task frame aligns with geometric features (surface normals, insertion axes) rather than contact dynamics, it is inherently sim-to-real transferable.
 
        We adapt the TFF in two ways. First, we extend the binary force/position split to three modes: force control, high-stiffness position control, and low-stiffness compliant position control. Free directions subject to resistance (friction, latches) receive high-stiffness control, while those without resistance receive low-stiffness compliant control. Constrained directions with a zero force set-point are also assigned low-stiffness compliant control, avoiding a failure mode where premature zero-setpoint force control eliminates positional correction before contact. Second, we fix the task frame origin at the end-effector, so only the frame \textit{orientation} needs to be predicted; this is valid because the manipulated object moves with the end-effector once contact is established.

        We encode the per-axis control modes in a six-dimensional control mode vector $C$. The encoding is fixed across all four tasks: 0 for force control, 1 for low-stiffness compliant position control, and 4 for high-stiffness position control. Taking whiteboard wiping as an example, the control mode vector is $C = [4, 4, 0, 1, 1, 4]^\top$ in the axis order $(x, y, z, r_x, r_y, r_z)$: Z translation is force-controlled to maintain board contact, XY translations and Z rotation are high-stiffness to overcome friction, and XY rotations are low-stiffness compliant to tolerate prediction noise and disturbances.

    \subsection{Privileged Expert Policy for Demonstration Collection} \label{sec:demo}

        We construct an expert policy in simulation that leverages ground-truth state, effortless resets and randomization, and parallel execution to generate large-scale demonstrations automatically.
        
        The expert policy organizes each task into two types of phases: \textit{free motion} and \textit{contact interaction}. In the free motion phase (pre-contact approach and post-task retraction), object-centric key poses are defined relative to task-relevant object parts, and trajectories are planned between the current and target poses. Since force regulation is unnecessary in this phase, the task frame is set to align with the world frame and the control mode vector is set to $G = [1, 1, 1, 1, 1, 1]^\top$ (low-stiffness compliant control along all axes). This enables smooth contact transitions without precise mode-switching.

\begin{figure}
        \centering
        \includegraphics[width=\textwidth]{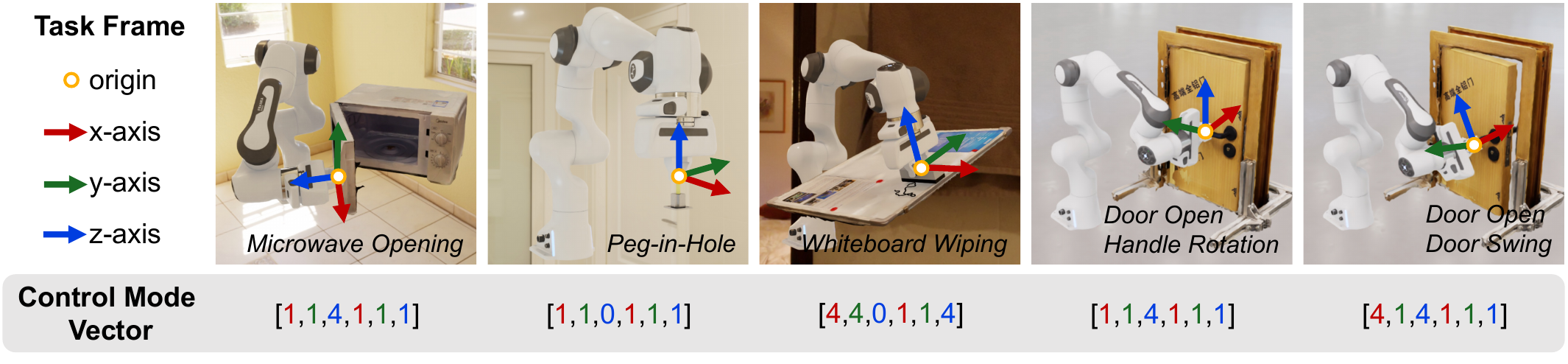}
        \caption{Ground-truth task frame and control mode vector.}
        \label{fig:task_frame}
    \end{figure}

        In the contact interaction phase, the expert policy executes task-specific motion rules designed to satisfy contact constraints: for peg-in-hole, the end-effector descends along the hole axis; for whiteboard wiping, a heuristic planner generates a zigzag coverage trajectory over the ink region; for articulated objects (microwave and door), an ideal circular trajectory about the joint axis is computed via Rodrigues' rotation formula using the joint axis and pivot from the simulator. The task frame orientation and control mode vector for each task are defined analytically from object geometry and summarized in Fig.~\ref{fig:task_frame}. For door opening, per-axis control modes differ between the handle rotation and door swing stages, with the transition triggered by thresholding the handle joint angle. The angle threshold is determined from measurements of the real door hardware.

        We address the sim-to-real gap along three axes. For the \textit{visual gap}, we apply domain randomization over backgrounds, lighting, and camera parameters atop photorealistic simulators~\cite{isaacsim, isaaclab, sapien, genesis}. For the \textit{geometric gap}, real-world objects are 3D-scanned and reconstructed as textured meshes imported into the simulator; articulated objects are segmented and assembled with joints. For the \textit{dynamics gap}, as discussed in Sec.~\ref{sec:taskframe} and Sec.~\ref{sec:policy}, we formulate the policy with dynamics-invariant inputs and outputs, eliminating the need for dynamics randomization entirely. Object poses and robot initial configurations are also randomized across episodes. Task completion is assessed via object poses or joint angles, retaining only successful episodes as demonstrations.

    \subsection{Visuomotor Policy} \label{sec:policy}

        As established in Sec.~\ref{sec:intro}, the directional component of force regulation is time-varying and impossible to specify manually, yet transferable from simulation. We train a visuomotor policy to predict it via imitation learning. The policy maps the current observation $O_t$ to a sequence of $h$ future actions $\{A_{t+k}\}_{k=0}^{h-1}$, where action chunking reduces compounding errors and improves temporal smoothness. The observation $O_t$ consists of: RGB images $I_t$ from multiple cameras (one wrist-mounted view and one to two fixed third-person views), proprioception $X_t \in \mathbb{R}^{10}$ containing end-effector position (3), 6D rotation~\cite{6drot} (6), and gripper state (1), and a task-level language instruction (e.g., ``open the microwave''). Each action $A_t$ comprises: end-effector pose and gripper command ($X_t^d \in \mathbb{R}^{10}$), task frame orientation ($T_t^d \in \mathbb{R}^{6}$), and control mode vector ($C_t^d \in \mathbb{R}^{6}$). We omit the subscript $t$ where no ambiguity arises.

        We instantiate the policy using E2VLA~\cite{e2vla}, a VLA model with strong 3D geometric reasoning capability, which aligns visual representations and action tokens with the underlying 3D physical space. To accommodate our expanded action space, we extend the input and output projection layers of E2VLA's diffusion action expert. The additional weights are initialized with small magnitudes following~\cite{tavla}, while original projection weights are inherited from the pre-trained checkpoint, preserving pre-trained behavior while allowing gradual learning of new targets.

        The training objective extends the original E2VLA loss, which supervises end-effector pose and gripper command, with additional terms for task frame orientation and control mode vector. Let $X^d_{\mathrm{pos}} \in \mathbb{R}^3$, $X^d_{\mathrm{rot}} \in \mathbb{R}^6$, and $X^d_{\mathrm{grip}} \in \mathbb{R}$ denote the position, 6D rotation, and gripper command components of $X^d$, respectively:
        \begin{equation}
            \mathcal{L} = \lambda_1 \Vert X^d_{\mathrm{pos}} - \hat{X}^d_{\mathrm{pos}} \Vert + \lambda_2 \Vert X^d_{\mathrm{rot}} - \hat{X}^d_{\mathrm{rot}} \Vert + \lambda_3 \Vert X^d_{\mathrm{grip}} - \hat{X}^d_{\mathrm{grip}} \Vert + \lambda_4 \Vert T^d - \hat{T}^d \Vert + \lambda_5 \Vert C^d - \hat{C}^d \Vert,
        \end{equation}
        where $\hat{\cdot}$ denotes ground-truth supervision targets and the loss weights are $(\lambda_1,\lambda_2,\lambda_3,\lambda_4,\lambda_5)=(30,30,10,30,30)$.

    \subsection{Adaptive Admittance Controller} \label{sec:controller}

        At deployment, the policy's predicted task frame orientation and control mode vector are consumed by a Cartesian admittance controller, whose per-axis stiffness and damping can be modulated. The controller dynamics are:
        \begin{equation}
            M\ddot{p}_r + D\dot{p}_r + K(p_r-p_{cmd}) = F_{ext} - F_{cmd},
        \end{equation}
        where $p_r$, $\dot{p}_r$, $\ddot{p}_r$ denote the compliant reference pose, twist, and acceleration, respectively. The commanded pose $p_{cmd}$ is the future end-effector pose predicted by the policy, and $F_{ext}$ is the external wrench acting on the end-effector. The mass matrix $M$ is diagonal with elements $m = 1$ for translational axes and $m = 0.1$ for rotational axes.

        The stiffness and damping matrices are computed in the task frame and then rotated to the world frame. Let $C^d \in \mathbb{R}^6$ be the control mode vector predicted by the policy. The task-frame stiffness matrix $\widetilde{K} \in \mathbb{R}^{6\times6}$ is diagonal, with entries
        \[
            \widetilde{K}_{ii} =
            \begin{cases}
                0, & C^d(i) = 0,\\
                k_{\mathrm{low}}, & C^d(i) = 1,\\
                k_{\mathrm{high}}, & C^d(i) = 4,
            \end{cases}
            \qquad i = 1,\ldots,6,
        \]
        where $k_{\mathrm{low}}$ and $k_{\mathrm{high}}$ are the manually specified low and high stiffnesses, with separate values for translational and rotational axes. The damping matrix is set as $\widetilde{D} = 2\xi\sqrt{M\widetilde{K}}$, where the square root is applied element-wise and $\xi = 2$ is an overdamped ratio that ensures controller stability. For force-controlled axes where $C^d(i) = 0$, both stiffness and damping vanish; we therefore set the damping for these axes to $2\xi\sqrt{m k_{\mathrm{low}}}$ to maintain stability. Letting $\mathcal{S} = \mathrm{diag}(R, R) \in \mathbb{R}^{6\times6}$ denote the block-diagonal matrix constructed from the task frame rotation matrix $R \in \mathbb{R}^{3\times3}$, the world-frame matrices are $K = \mathcal{S}\widetilde{K}\mathcal{S}^\top$ and $D = \mathcal{S}\widetilde{D}\mathcal{S}^\top$.

        In the task frame, the commanded wrench is nonzero only along force-controlled axes:
        \begin{equation}
            \widetilde{F}_{cmd}(i) = F_h(i) \cdot \mathbf{1}[C^d(i) = 0], \quad i = 1, \ldots, 6,
        \end{equation}
        where $F_h \in \mathbb{R}^6$ is the manually specified target wrench and $\mathbf{1}[\cdot]$ is the indicator function. The world-frame commanded wrench is $F_{cmd} = \mathcal{S}\widetilde{F}_{cmd}$.

        The resulting control behavior follows from the control mode along each task frame axis: 0 yields force control (zero stiffness, driven by $F_{cmd}$), 1 yields low-stiffness compliant position control ($k_{\mathrm{low}}$), and 4 yields high-stiffness position control ($k_{\mathrm{high}}$). In our default experimental setting, we fix $k_{\mathrm{high}} = 4k_{\mathrm{low}}$, so only $k_{\mathrm{low}}$ needs to be specified for stiffness. We denote it by $k$, giving $k_{\mathrm{high}} = 4k$. The two stiffnesses can also be specified independently at deployment without changing the control mode encoding. The target force magnitudes $F_h$ along force-controlled axes are specified separately.


\section{Experiment}

    \begin{table}[t]
        \centering
        \resizebox{\textwidth}{!}{%
            \begin{tabular}{@{}lcccccccccc@{}}
                \toprule
                 &
                  \multicolumn{2}{c}{MO} &
                  \multicolumn{2}{c}{PH} &
                  \multicolumn{2}{c}{WW} &
                  \multicolumn{2}{c}{DO} &
                  \multicolumn{2}{c}{Overall} \\ \cmidrule(l){2-3} \cmidrule(l){4-5} \cmidrule(l){6-7} \cmidrule(l){8-9} \cmidrule(l){10-11}
                              & SR-ND & SR-D & SR-ND & SR-D                        & SR-ND                       & SR-D & SR-ND & SR-D                        & SR-ND & SR-D \\ \midrule
                Pi0-Real      & -     & -    & 30    & 40                          & 60                          & 20   & -     & -                           & 45    & 30   \\
                E2VLA-Real    & -     & -    & 35    & 20                          & 90                          & 60   & -     & -                           & 63    & 40   \\
                ForceVLA-Real & -     & -    & 15    & 40                          & \cellcolor[HTML]{E1F0D6}100 & 80   & -     & -                           & 58    & 60   \\
                ACP-Real      & -     & -    & 45    & 20                          & 60                          & 0    & -     & -                           & 53    & 10   \\ \midrule
                Pi0-Sim+L-A   & 30    & 60   & 35    & 60                          & 35                          & 60   & 80    & \cellcolor[HTML]{E1F0D6}100 & 45    & 70   \\
                Pi0-Sim+M-A   & 75    & 60   & 30    & 60                          & 60                          & 60   & 90    & 40                          & 64    & 55   \\
                Pi0-Sim+H-A   & 60    & 40   & 35    & 20                          & 45                          & 20   & 80    & 40                          & 55    & 30   \\
                E2VLA-Sim+L-A & 15    & 40   & 55    & \cellcolor[HTML]{E1F0D6}100 & 30                          & 60   & 65    & 80                          & 41    & 70   \\
                E2VLA-Sim+M-A & 90    & 20   & 55    & 60                          & 55                          & 80   & 85    & 40                          & 71    & 50   \\
                E2VLA-Sim+H-A & 70    & 20   & 70    & 40                          & 65                          & 40   & 75    & 40                          & 70    & 35   \\
                ForceVLA-Sim  & 0     & 0    & 0     & 60                          & 5                           & 40   & 0     & 20                          & 1     & 30   \\
                ACP-Sim       & 0     & 0    & 5     & 20                          & 15                          & 0    & 0     & 0                           & 5     & 5    \\ \midrule
                Ours &
                  \cellcolor[HTML]{E1F0D6}95 &
                  \cellcolor[HTML]{E1F0D6}80 &
                  \cellcolor[HTML]{E1F0D6}90 &
                  \cellcolor[HTML]{E1F0D6}100 &
                  \cellcolor[HTML]{E1F0D6}100 &
                  \cellcolor[HTML]{E1F0D6}100 &
                  \cellcolor[HTML]{E1F0D6}95 &
                  80 &
                  \cellcolor[HTML]{E1F0D6}95 &
                  \cellcolor[HTML]{E1F0D6}90 \\ \bottomrule
            \end{tabular}%
        }
        \caption{Success rate (sr) comparison. SR-ND: sr without disturbance; SR-D: sr with disturbance. ``-'' indicates an unavailable real-data baseline, not a zero success rate.}
        \vspace{-0.6cm}
        \label{tab:sr}
    \end{table} 

    We conduct experiments to answer the following questions: \textbf{(Q1)} Does TDC outperform a broad set of baselines across diverse contact-rich tasks? \textbf{(Q2)} How much does contact dynamics gap impact force-aware policies? \textbf{(Q3)} Does TDC benefit from large-scale pretraining and adaptive compliance? \textbf{(Q4)} How sensitive is TDC to the choice of stiffness and target force magnitude?

    \subsection{Setup} \label{sec:setup}

        \textbf{Platform.} The robot platform comprises a Franka arm with a Franka Hand gripper. Two Intel RealSense D435 cameras provide RGB images as a global camera and a wrist-mounted camera, respectively; the whiteboard wiping task uses an additional global camera to better capture the board pose. The external wrench on the end-effector is estimated from joint torque measurements and the robot dynamics model. All models run on a server with an NVIDIA RTX 4090.

        \textbf{Tasks.} As shown in Tab.~\ref{tab:task}, we evaluate on four contact-rich manipulation tasks: Microwave Opening (MO), Peg-in-Hole (PH), Whiteboard Wiping (WW), and Door Opening (DO), covering diverse interaction constraints and force requirements. The peg-in-hole task uses a peg-hole pair with 1 mm clearance and a full insertion depth of 25 mm.

        \textbf{Baselines.} We compare TDC against multiple baselines detailed in Tab.~\ref{tab:baseline}, which vary in architecture, data source, training strategy, input/output modality, and controller. ``L/M/H-A'' denotes admittance controllers with translational stiffness of 50, 200, and 800, respectively, and a uniform rotational stiffness of 10. For peg-in-hole and whiteboard wiping, we collect 100 real-world demonstrations each using the Touch teleoperation system with a mid-stiffness impedance controller, totaling about 8 person-hours. These demonstrations are used only for the ``-Real'' baselines; TDC does not use them. For microwave opening and door opening, the unintuitive force feedback of teleoperation almost always triggered safety emergency stops, preventing real-world data collection for these two tasks. We collect 2000 simulation demonstrations per task using the privileged expert policy (Sec.~\ref{sec:demo}), requiring approximately 6 hours per task on a single NVIDIA RTX 4090. For TDC, the stiffness $k$ is set to 50/10 (translational/rotational) across all tasks. The target force magnitude is set to 2N (PH) or 4N (WW) along the translational z-axis of task frame and zero elsewhere; MO and DO require no target force.

        \textbf{Evaluation.} For each task and method, we execute 20 rollouts with randomized object and robot initial poses. We additionally apply disturbances in 5 trials that begin at the contact state (grasping handles, partial peg insertion, eraser-whiteboard engagement). Disturbances are introduced by having a human operator apply external forces to the manipulated object during the contact interaction phase, perturbing its pose. A trial is successful if the robot completes the task within the time limit without triggering safety stops. The primary metric is success rate (\%). For PH and WW, we also report task completion quality: insertion depth (mm) and remaining ink length (cm), respectively.

    \begin{table}[t]
        \centering
        
        \begin{minipage}[t]{0.42\textwidth}
            \vspace{0pt}

            \begin{minipage}[t]{\linewidth}
                \centering
            
                \resizebox{\linewidth}{!}{
                    \begin{tabular}{@{}lcc@{}}
                        \toprule
                         & \begin{tabular}[c]{@{}c@{}}PH Insertion\\ Depth (mm) $\uparrow$\end{tabular} &
                         \begin{tabular}[c]{@{}c@{}}WW Remaining\\ Ink Length (cm) $\downarrow$\end{tabular} \\ 
                        \midrule
                        
                        Pi0-Real      & 23.1 & 1.7 \\
                        E2VLA-Real    & \cellcolor[HTML]{E1F0D6}25.0 & 0.8 \\
                        ForceVLA-Real & 22.0 & 0.5 \\
                        ACP-Real      & 21.0 & 1.3 \\
                        
                        \midrule
                        
                        Pi0-Sim+L-A   & 21.6 & 2.3 \\
                        Pi0-Sim+M-A   & 21.3 & 2.2 \\
                        Pi0-Sim+H-A   & 22.7 & 1.3 \\
                        E2VLA-Sim+L-A & 21.9 & 2.0 \\
                        E2VLA-Sim+M-A & 23.5 & 1.1 \\
                        E2VLA-Sim+H-A & 21.0 & 0.8 \\
                        ForceVLA-Sim  & 20.0 & 4.0 \\
                        ACP-Sim       & 17.5 & 3.7 \\
                        
                        \midrule
                        
                        Ours          & \cellcolor[HTML]{E1F0D6}25.0 &
                        \cellcolor[HTML]{E1F0D6}0.3 \\
                        
                        \bottomrule
                    \end{tabular}
                }
                \caption{Task completion quality, averaged over successful trials.}
                \label{tab:quality}

            \end{minipage} 


            \begin{minipage}[t]{\linewidth}
                \centering
                
                \resizebox{\linewidth}{!}{
                    \begin{tabular}{@{}lcccc@{}}
                        \toprule
                        Task &
                        \begin{tabular}[c]{@{}c@{}}Constraint\\ Switching\end{tabular} &
                        \begin{tabular}[c]{@{}c@{}}Resistance\\ Overcoming\end{tabular} &
                        \begin{tabular}[c]{@{}c@{}}Contact\\ Maintenance\end{tabular} &
                        \begin{tabular}[c]{@{}c@{}}High\\ Precision\end{tabular} \\
                        
                        \midrule
                        
                        MO & $\times$     & $\checkmark$ & $\times$     & $\times$     \\
                        PH & $\times$     & $\times$     & $\checkmark$ & $\checkmark$ \\
                        WW & $\times$     & $\checkmark$ & $\checkmark$ & $\times$     \\
                        DO & $\checkmark$ & $\checkmark$ & $\times$     & $\times$     \\
                        
                        \bottomrule
                    \end{tabular}
                }
                \caption{Task characteristics.}
                \label{tab:task}
            
            \end{minipage} 
        
        \end{minipage} 
        \hfill
        \begin{minipage}[t]{0.56\textwidth}
            \vspace{0pt}
            
            \begin{minipage}[t]{\linewidth}
                \centering
                
                \resizebox{\linewidth}{!}{
                    \begin{tabular}{@{}lccccc@{}}
                        \toprule
                        Baseline &
                          \begin{tabular}[c]{@{}c@{}}Real\\ Data\end{tabular} &
                          \begin{tabular}[c]{@{}c@{}}Pre-\\ Train\end{tabular} &
                          \begin{tabular}[c]{@{}c@{}}Force\\ In\end{tabular} &
                          \begin{tabular}[c]{@{}c@{}}Force\\ Out\end{tabular} &
                          Controller \\ \midrule
                        Pi0-Real~\cite{pi0}          & $\checkmark$ & $\checkmark$ & $\times$     & $\times$     & teleoperation impedance            \\
                        E2VLA-Real~\cite{e2vla}        & $\checkmark$ & $\checkmark$ & $\times$     & $\times$     & teleoperation impedance            \\
                        ForceVLA-Real~\cite{forcevla}     & $\checkmark$ & $\checkmark$ & $\checkmark$ & $\times$     & teleoperation impedance            \\
                        ACP-Real~\cite{acp}          & $\checkmark$ & $\times$     & $\checkmark$ & $\checkmark$ & ACP impedance                      \\ \midrule
                        Pi0-Sim+L/M/H-A~\cite{pi0}   & $\times$     & $\checkmark$ & $\times$     & $\times$     & low-/mid-/high-stiffness admittance \\
                        E2VLA-Sim+L/M/H-A~\cite{e2vla} & $\times$     & $\checkmark$ & $\times$     & $\times$     & low-/mid-/high-stiffness admittance \\
                        ForceVLA-Sim~\cite{forcevla}      & $\times$     & $\checkmark$ & $\checkmark$ & $\times$     & teleoperation impedance            \\
                        ACP-Sim~\cite{acp}           & $\times$     & $\times$     & $\checkmark$ & $\checkmark$ & ACP impedance                      \\ \bottomrule
                    \end{tabular}
                }
                \caption{Details of baselines. ``Force In'' and ``Force Out'' indicate whether the method incorporates force measurements as input or predicts force-related outputs, respectively.}
                \label{tab:baseline}
            
            \end{minipage} 
            
            
            \begin{minipage}[t]{\linewidth}
                \centering
                
                \resizebox{\linewidth}{!}{
                    \begin{tabular}{@{}ccccc@{}}
                        \toprule
                        No. &
                        \begin{tabular}[c]{@{}c@{}}Pre-\\ Train\end{tabular} &
                        \begin{tabular}[c]{@{}c@{}}Adaptive\\ Controller\end{tabular} &
                        \begin{tabular}[c]{@{}c@{}}SR-ND\\ (\%) $\uparrow$\end{tabular} &
                        \begin{tabular}[c]{@{}c@{}}SR-D\\ (\%) $\uparrow$\end{tabular}\\
                        
                        \midrule
                        
                        1 & $\times$     & $\checkmark$ & 76.25 [65.86, 84.24] &
                        \cellcolor[HTML]{E1F0D6}95.00 [76.39, 99.11] \\
                        
                        2 & $\checkmark$ & L-A & 36.25 [26.57, 47.19] & 60.00 [38.66, 78.12] \\
                        3 & $\checkmark$ & M-A & 65.00 [54.08, 74.55] & 40.00 [21.88, 61.34] \\
                        4 & $\checkmark$ & H-A & 51.25 [40.49, 61.89] & 10.00 [2.79, 30.10]  \\
                        
                        \midrule
                        
                        5 & $\checkmark$ & $\checkmark$ &
                        \cellcolor[HTML]{E1F0D6}95.00 [87.84, 98.04] &
                        90.00 [69.90, 97.21] \\
                        
                        \bottomrule
                    \end{tabular}
                }
                \caption{Ablation results pooled across all four tasks. Each row includes 80 trials for SR-ND and 20 for SR-D. Brackets indicate 95\% Wilson confidence intervals.}
                \label{tab:ablation}
            
            \end{minipage} 
        
        \end{minipage}

    \vspace{-0.8cm}
    \end{table} 

    \subsection{Results} \label{sec:results}

\begin{figure}[t]

        \centering
    
        \begin{minipage}[t]{0.49\textwidth}
            \centering
            \includegraphics[width=\linewidth]{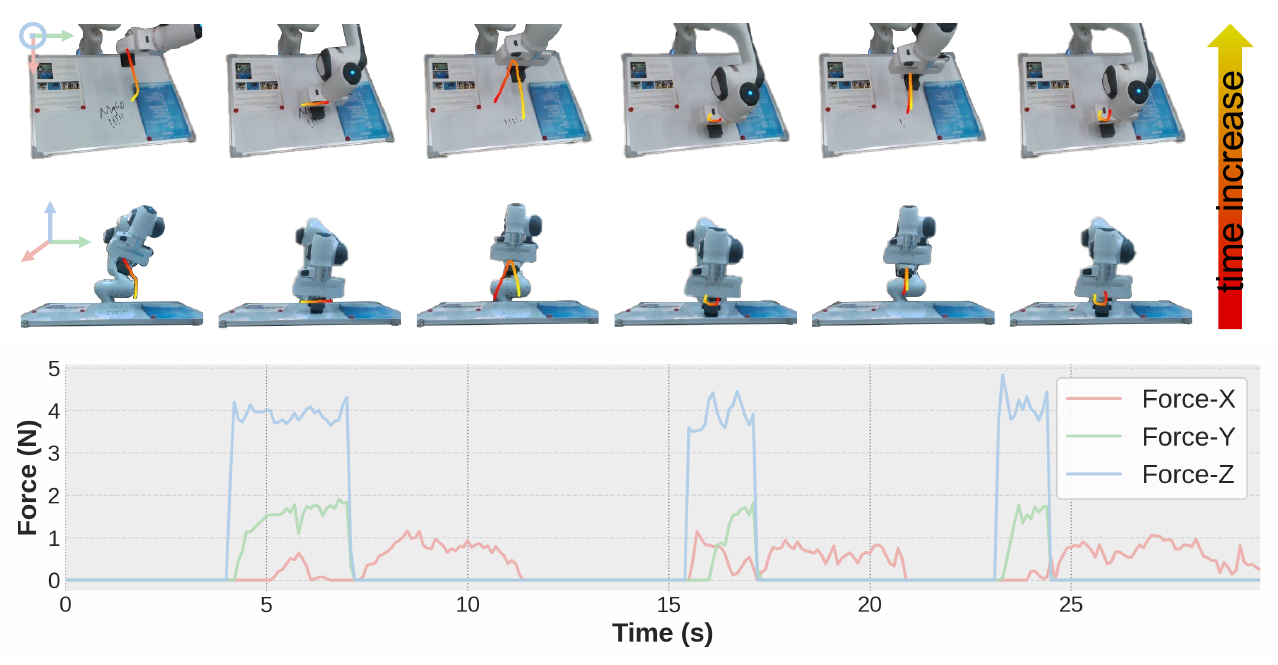}
            \captionof{figure}{Force variation of TDC during wiping.}
            \vspace{-0.5cm}
            \label{fig:force}
        \end{minipage}
        \hfill
        \begin{minipage}[t]{0.49\textwidth}
            \centering
            \includegraphics[width=\linewidth]{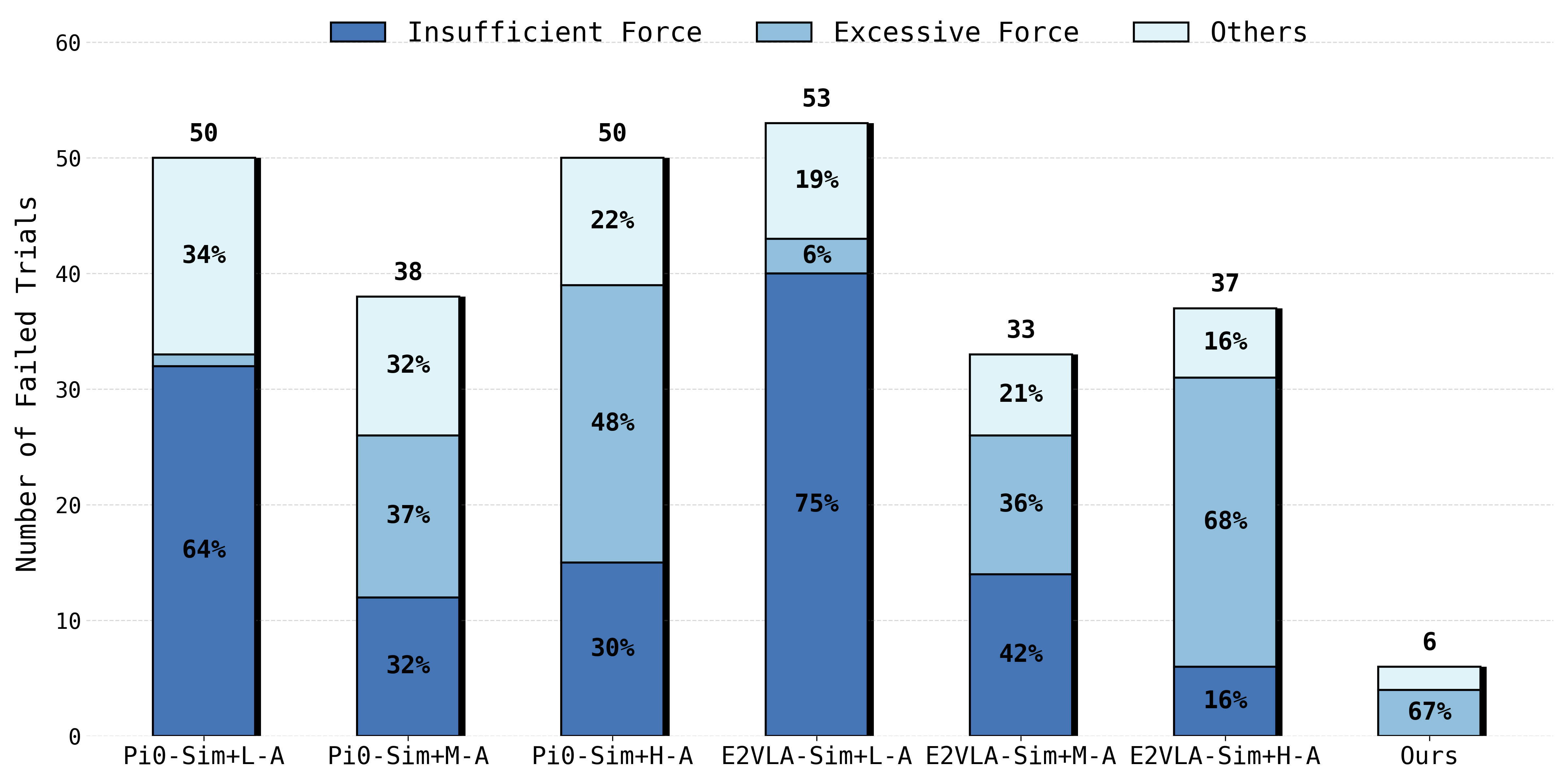}
            \vspace{-0.38cm}
            \captionof{figure}{Failure mode statistics.}
            \label{fig:failure}
        \end{minipage}
        \vspace{-0.1cm}
    \end{figure}

        \textbf{TDC consistently outperforms all baselines across diverse contact-rich tasks (Q1).} Tab.~\ref{tab:sr} summarizes success rates under undisturbed (SR-ND) and disturbed (SR-D) conditions. Our method improves overall SR-ND and SR-D over the best baseline by 24 and 20 percentage points, respectively. On peg-in-hole and whiteboard wiping, TDC outperforms baselines trained on real-world data in both effectiveness and robustness. Note that higher SR-D than SR-ND in some models results from the evaluation protocol (Sec.~\ref{sec:setup}), where disturbed trials begin at the contact state. Tab.~\ref{tab:quality} shows that TDC also achieves higher task completion quality, i.e., deeper insertions and more thorough wiping. Fig.~\ref{fig:force} shows the force profile during wiping: the normal contact force (Force-Z) converges rapidly to the 4N target upon contact.

        \textbf{Contact dynamics gap poses a significant challenge for force-aware policies trained in simulation (Q2).} As shown in Tab.~\ref{tab:sr} and Tab.~\ref{tab:quality}, force-aware baselines trained on simulation data (ForceVLA-Sim and ACP-Sim) achieve consistently low success rates and task completion quality, with a substantial performance drop compared to their real-world-trained counterparts.

        \textbf{Both large-scale pretraining and adaptive compliance contribute to TDC's performance (Q3).} We report ablation results pooled across all four tasks (Tab.~\ref{tab:ablation}). Finetuning from a pre-trained VLA (No.5) improves the aggregate SR-ND from 76.25\% to 95.00\% compared with training from scratch (No.1), confirming the effectiveness of the architecture and training strategy in Sec.~\ref{sec:policy}. Despite lacking pre-training, No.1 outperforms No.2-4, which use the same policy as No.5 but replace the adaptive controller with isotropic admittance controllers. This highlights that force regulation plays a critical role in contact-rich tasks, and large-scale position-only pre-training alone is insufficient. The improvement of No.5 over No.2-4 validates our adaptive compliance. We further categorize failures into three types: insufficient force, excessive force, and others. Excessive-force failures are recorded automatically by safety stops; an operator distinguishes insufficient force from other errors using the criteria in Appendix~\ref{sec:evaluation_protocols}. As shown in Fig.~\ref{fig:failure}, baselines with isotropic admittance controllers are dominated by force-related failures, while TDC substantially reduces those failures. TDC fails in 6 of 100 trials: four excessive-force failures caused by occasional task frame misalignment and two other errors caused by pose prediction errors.

\begin{figure}
        \centering
        \includegraphics[width=\textwidth]{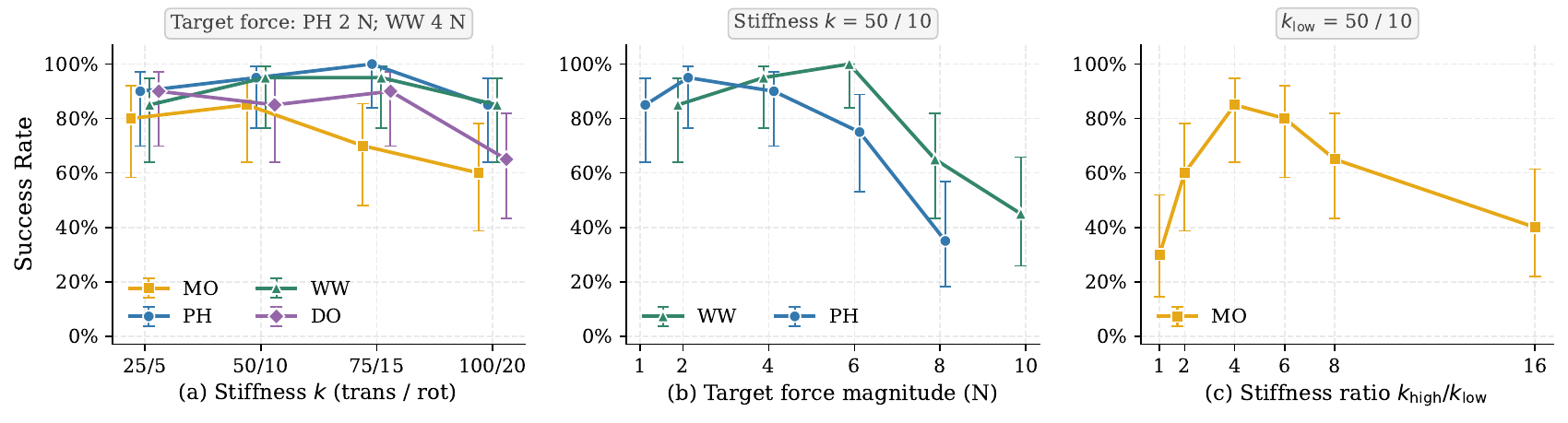}
        \vspace{-0.5cm}
        \caption{Sensitivity analysis: (a) stiffness $k$ across four tasks; (b) target force magnitude on WW and PH; (c) high-to-low stiffness ratio on MO with $k_{\mathrm{low}}$ set to 50/10. In (a) and (b), task curves and error bars are slightly offset horizontally for visibility; ticks indicate the actual parameter values.}
        \label{fig:sensitivity}
        \vspace{-0.3cm}
    \end{figure}

        \textbf{TDC shows low sensitivity within broad parameter ranges (Q4).} 
        We evaluate sensitivity to stiffness $k$ across all four tasks, target force magnitude on WW and PH, and the high-to-low stiffness ratio on MO (Fig.~\ref{fig:sensitivity}). Each point pools 10 undisturbed and 10 disturbed trials, with 95\% Wilson confidence intervals. The stiffness $k$ sweep uses $k_{\mathrm{high}}=4k$ and the task-specific target forces from Sec.~\ref{sec:setup}; the force sweep fixes $k$ at 50/10 (translational/rotational). For the ratio sweep, $k_{\mathrm{low}}$ is 50/10 (translational/rotational), and the ratio $k_{\mathrm{high}}/k_{\mathrm{low}}$ ranges from 1 to 16. Success rates remain high over broad ranges of stiffness $k$ and target force magnitude, but decline at large stiffness or force values. On MO, equal low and high stiffnesses yield 30\% success, compared with 85\% at a ratio of 4, while increasing the ratio to 16 reduces success to 40\%. These results support low sensitivity over broad usable ranges, with performance declining at extreme values.

\section{Conclusion} \label{sec:Conclusion}

    We presented \textbf{TDC}, a sim-to-real framework for adaptive compliance in contact-rich manipulation that requires no real-world demonstrations. Our core idea is to decompose force regulation into a transferable directional component, predicted by a VLA-based visuomotor policy trained on simulation demonstrations, and a manually tunable magnitude component specified at deployment. Experiments on four contact-rich tasks demonstrate strong success rates and robustness to external disturbances.


\section{Limitations} \label{sec:limitations}

    TDC has several limitations. First, fixing the task frame origin at the end-effector assumes the contact point coincides with it, which breaks down for tasks where contact is spatially decoupled from the end-effector and torque regulation at that point is required. Second, expanding the action space with new prediction targets degrades performance on the original targets (cf.\ No.2-4 in Tab.~\ref{tab:ablation} vs.\ E2VLA-Sim+L/M/H-A in Tab.~\ref{tab:sr}), revealing a tension between preserving pre-trained capabilities and learning new targets that our selective initialization only partially mitigates. Third, manually authored planners have limited scalability to tasks with complex contact patterns. Fourth, using constant stiffness and target force magnitudes throughout execution requires the feasible parameter ranges of different task stages to overlap. Finally, TDC has not been evaluated on tasks involving multiple simultaneous contacts, such as dexterous hands.


\clearpage
\acknowledgments{This work was supported by the Fundamental and Interdisciplinary Disciplines Breakthrough Plan of the Ministry of Education of China under Grant No. JYB2025XDXM208, the National Nature Science Foundation of China under Grant No. 62522317, and Zhejiang Provincial Natural Science Foundation of China under Grant No. LD25F030001.}


\bibliography{example}  

\clearpage

\appendix

\section{Details on Domain Randomization in Simulation}

    All randomization parameters are sampled independently at the start of each episode. We divide them into shared parameters (Tab.~\ref{tab:shared_rand}), which are identical across all four tasks, and task-specific parameters (Tab.~\ref{tab:task_rand}), which vary by task. For background randomization, we use a pool of 27 HDR background images and sample uniformly. For camera intrinsic randomization, a single offset is drawn from $\mathcal{U}(-10, 10)$ and added to both $f_x$ and $f_y$; the principal point ($c_x$, $c_y$) is kept at its default value. The whiteboard wiping task uses a second global camera whose randomization is identical to the shared global camera. Ink strokes on the whiteboard are procedurally generated with randomized shapes and placed at random positions on the board surface.

    \begin{table}[h]
        \centering
        \begin{tabular}{@{}lll@{}}
            \toprule
                                           & Parameter           & Distribution                \\ \midrule
            Background                     & HDR image           & $\mathcal{U}\{1, ..., 27\}$ \\ \midrule
            \multirow{2}{*}{Lighting}      & light intensity     & $\mathcal{N}(1000, 200^2)$  \\
                                           & color temperature   & $\mathcal{N}(6500, 500^2)$  \\ \midrule
            \multirow{7}{*}{Global Camera} & focal length offset & $\mathcal{U}(-10, 10)$      \\
                                           & x offset (m)        & $\mathcal{U}(-0.02, 0.02)$  \\
                                           & y offset (m)        & $\mathcal{U}(-0.02, 0.02)$  \\
                                           & z offset (m)        & $\mathcal{U}(-0.02, 0.02)$  \\
                                           & roll offset (deg)   & $\mathcal{U}(-5, 5)$    \\
                                           & pitch offset (deg)  & $\mathcal{U}(-5, 5)$    \\
                                           & yaw offset (deg)    & $\mathcal{U}(-5, 5)$    \\ \midrule
            \multirow{7}{*}{Wrist Camera}  & focal length offset & $\mathcal{U}(-10, 10)$      \\
                                           & x offset (m)        & $\mathcal{U}(-0.02, 0.02)$  \\
                                           & y offset (m)        & $\mathcal{U}(-0.02, 0.02)$  \\
                                           & z offset (m)        & $\mathcal{U}(-0.02, 0.0)$   \\
                                           & roll offset (deg)   & $\mathcal{U}(-5, 5)$    \\
                                           & pitch offset (deg)  & $\mathcal{U}(-5, 5)$    \\
                                           & yaw offset (deg)    & $\mathcal{U}(-5, 5)$    \\ \bottomrule
        \end{tabular}
        \caption{Shared domain randomization parameters applied to all four tasks. $\mathcal{U}$ and $\mathcal{N}$ denote uniform and Gaussian distributions, respectively.}
        \label{tab:shared_rand}
    \end{table}

    \begin{table}[h]
        \centering
        \resizebox{\textwidth}{!}{
            \begin{tabular}{@{}llcccccc@{}}
                \toprule
                Task & Parameter             & x (m)         & y (m)         & z (m)           & roll (deg)  & pitch (deg) & yaw (deg)   \\ \midrule
                \multirow{2}{*}{MO} & microwave pose  & (-0.1, 0.1) & (-0.2, 0.2)   & (-0.1, 0.1) & ---         & --- & (-45, 45) \\
                     & robot initial pose    & (-0.2, 0.2)   & (-0.3, 0.3)   & (-0.2, 0.2)     & (-45, 45) & (-45, 45) & (-45, 45) \\ \midrule
                \multirow{3}{*}{PH}       & hole pose       & (-0.2, 0.2)   & (-0.2, 0.2)   & (-0.1, 0.1)   & ---         & --- & (-45, 45) \\
                     & peg grasp pose        & ---           & ---           & (-0.005, 0.005) & ---         & ---         & ---         \\
                     & robot initial pose    & (-0.2, 0.2)   & (-0.3, 0.3)   & (-0.3, 0.2)     & (-45, 45) & (-45, 45) & (-45, 45) \\ \midrule
                \multirow{4}{*}{WW}                  & whiteboard pose & (-0.05, 0.05) & (-0.05, 0.05) & (-0.1, 0.1)  & (-20, 20) & --- & (-15, 15) \\
                     & ink position on board & (-0.15, 0.15) & (-0.28, 0.28) & ---             & ---         & ---         & ---         \\
                     & eraser grasp pose     & (-0.01, 0.01) & ---           & (-0.005, 0.005) & ---         & (-5, 5) & ---         \\
                     & robot initial pose    & (-0.1, 0.1)   & (-0.1, 0.1)   & (-0.1, 0.1)     & (-30, 30) & (-30, 30) & (-30, 30) \\ \midrule
                \multirow{2}{*}{DO}      & door pose       & (-0.05, 0.05) & (-0.2, 0.2) & (-0.1, 0.1)   & ---         & --- & (-15, 15) \\
                     & robot initial pose    & (-0.2, 0.2)   & (-0.3, 0.3)   & (-0.3, 0.1)     & (-45, 45) & (-45, 45) & (-45, 45) \\ \bottomrule
            \end{tabular}
        }
        \caption{Task-specific randomization parameters. All offsets are uniformly sampled and applied to default values. ``---'' indicates no randomization along that axis.}
        \label{tab:task_rand}
    \end{table}

\section{Per-Task Human Effort}

    Setting up the simulation environment and authoring the task-specific planner take approximately 1 and 5 person-hours per task, respectively. Phase splitting, task frame labeling, and control mode vector labeling each take less than 5 minutes per task. This one-off effort enables large-scale generation of high-quality demonstrations, with negligible additional human effort for subsequent data generation for the same task.

\section{Implementation Details}

    All training hyperparameters of TDC follow the original implementation of E2VLA. Training TDC on a single NVIDIA RTX 4090 takes approximately 10 hours. All baselines are trained with the hyperparameters specified in their original papers. All methods use RGB images at $240 \times 320$ resolution. ACP uses observation history including 2 RGB frames, 3 end-effector pose and gripper state frames, and 32 end-effector wrench frames. All other methods observe only the current timestep. The action chunk size follows each method's original setting: 16 for ACP, 32 for E2VLA and TDC, and 50 for ForceVLA and Pi0.  

\section{Hardware Settings}

\begin{figure}[h]
        \centering
        \includegraphics[width=\textwidth]{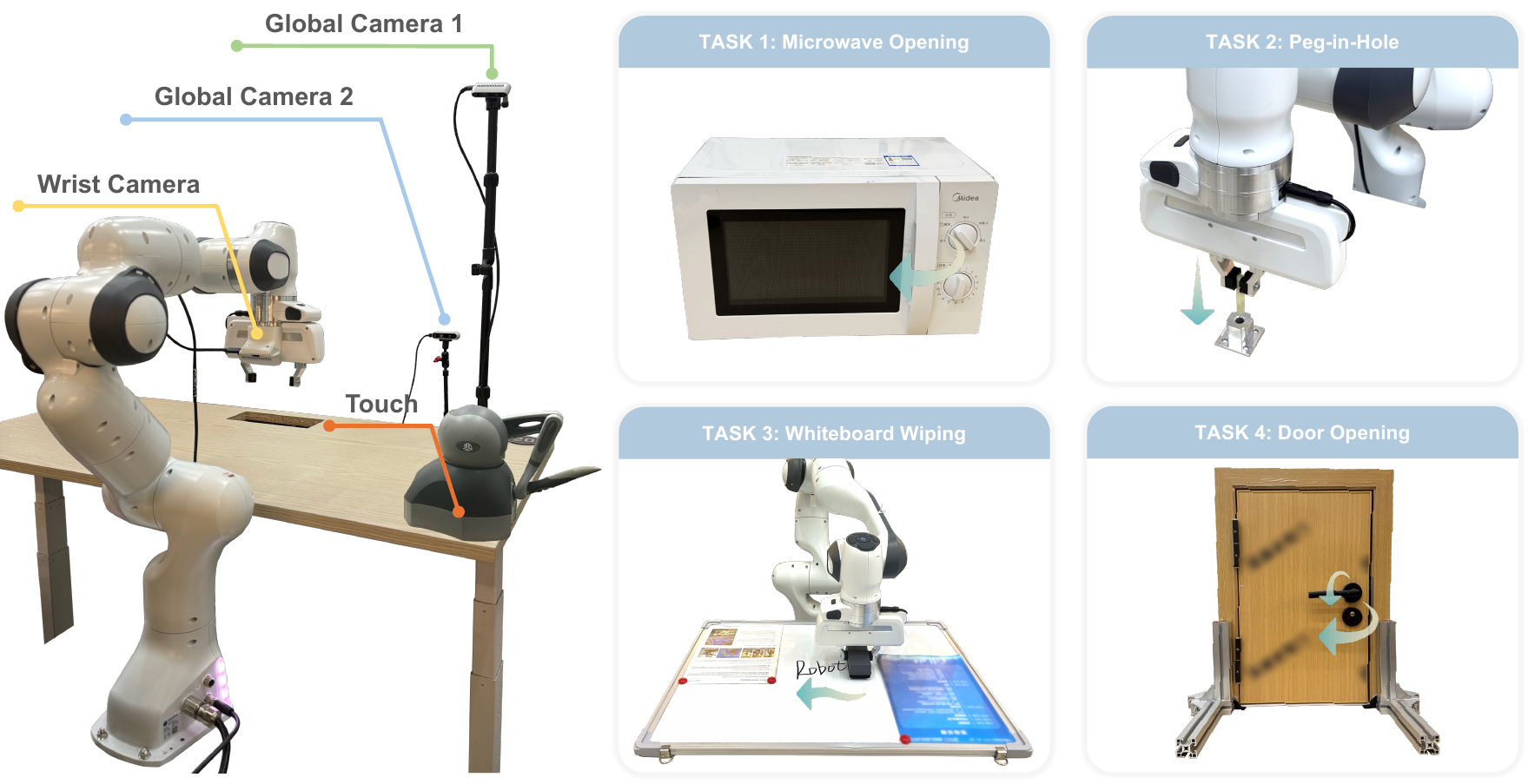}
        \caption{The real-world hardware settings.}
        \label{fig:platform}
    \end{figure}

    Fig.~\ref{fig:platform} shows our real-world experimental platform and objects involved in each task.

\section{Details on Evaluation Protocols} \label{sec:evaluation_protocols}

    The object pose randomization ranges, success criteria, and disturbance protocols for each task are detailed below. As stated in Sec.~\ref{sec:setup}, a trial is considered successful if the robot completes the task-specific criterion within the time limit without triggering safety stops.
    
    \textbf{Microwave Opening.} The microwave is randomly placed on a table within a 10\,cm $\times$ 20\,cm $\times$ 10\,cm area, with random orientations around the vertical axis within $\pm 15^{\circ}$. Success requires the door to be opened to at least $50^{\circ}$ within 120 seconds. Disturbances are introduced by randomly displacing and reorienting the microwave body during the door-opening process.
    
    \textbf{Peg-in-Hole.} The hole is randomly placed on a table within a 40\,cm $\times$ 40\,cm $\times$ 20\,cm area, with random orientations around the vertical axis within $\pm 45^{\circ}$. The hole depth is 25\,mm, and success requires the peg to be inserted to at least 10\,mm within 60 seconds. Disturbances are introduced after partial insertion by raising, lowering, or horizontally displacing the hole.
    
    \textbf{Whiteboard Wiping.} The whiteboard is placed at a fixed location on a table, with random height adjustments within 10\,cm and random lateral tilt within $\pm 20^{\circ}$. Before each trial, the human evaluator writes random scribbles on the board, keeping their shape and location similar across all methods. Success requires the remaining ink length to be less than 5\,cm within 120 seconds. Disturbances are introduced during wiping by raising, lowering, tilting, or horizontally displacing the whiteboard.
    
    \textbf{Door Opening.} The door is randomly placed within a 10\,cm $\times$ 30\,cm $\times$ 10\,cm area, with random orientations around the vertical axis within $\pm 15^{\circ}$. Success requires the door to be opened to at least $30^{\circ}$ within 120 seconds. Disturbances are introduced by randomly displacing and reorienting the door during the handle-turning and door-swinging phases.

    \textbf{Failure classification.} Excessive-force failures are recorded automatically when a safety stop is triggered. For the remaining failures, an operator assesses whether the intended contact is established and the motion direction or trajectory is essentially correct. A failure is classified as insufficient force when these conditions hold but the robot cannot complete the required interaction: in MO, the robot grasps the handle but cannot open the microwave door; in PH, the peg does not reach sufficient insertion depth; in WW, the wiping trajectory covers the ink but fails to remove it; and in DO, the robot grasps the handle but cannot rotate it sufficiently to unlock the door or cannot open the door to the required angle. Failures due to incorrect motion trajectory, geometric misalignment, or unsuccessful grasping are classified as others.

\section{Additional Qualitative Analysis of Force Regulation}

    \begin{figure}[h]
        \centering

        \begin{subfigure}{0.48\textwidth}
            \centering
            \includegraphics[width=\linewidth]{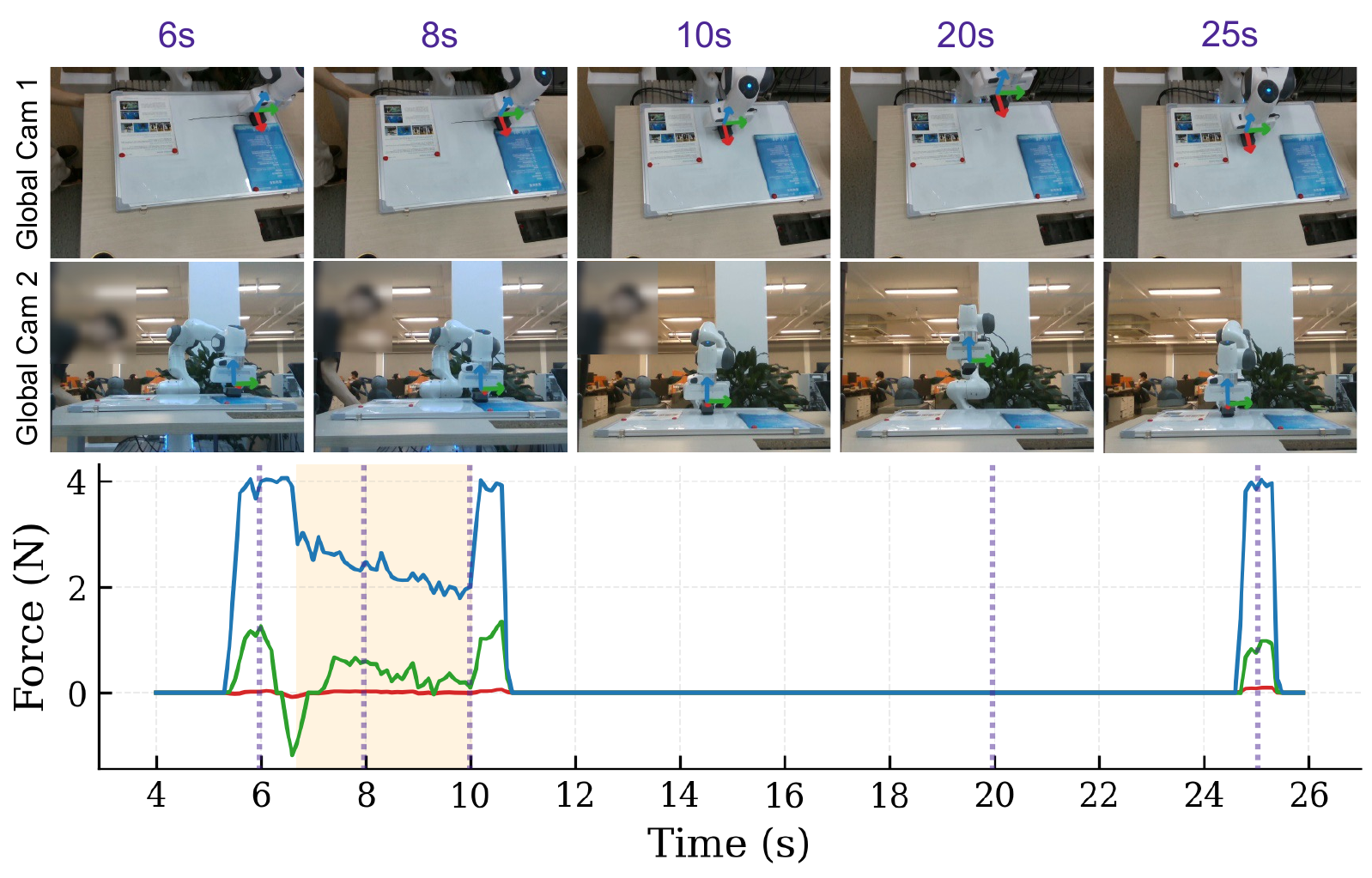}
            \vspace{-0.5cm}
            \caption{Whiteboard Wiping: board lowered.}
        \end{subfigure}
        \hfill
        \begin{subfigure}{0.48\textwidth}
            \centering
            \includegraphics[width=\linewidth]{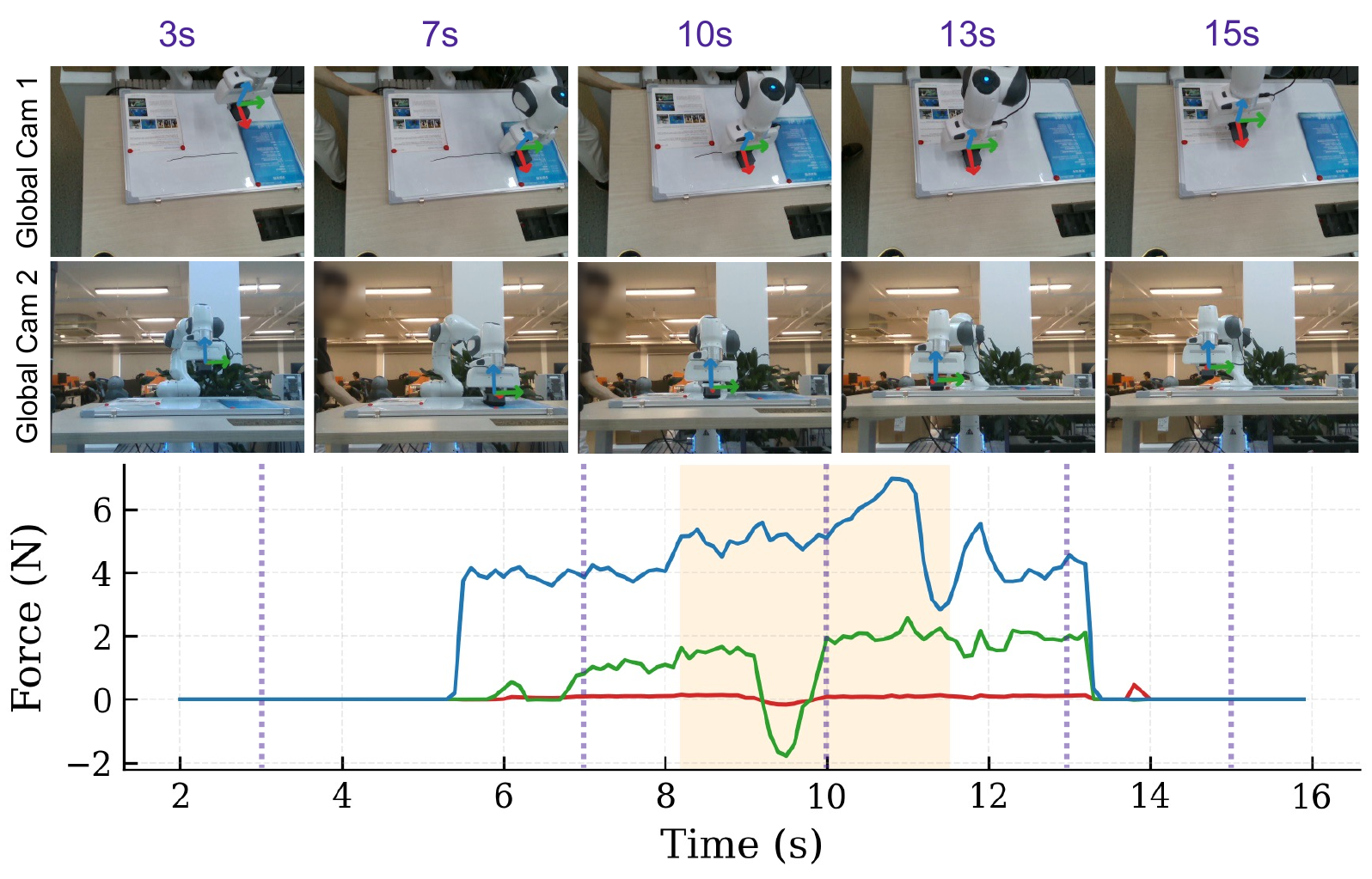}
            \vspace{-0.5cm}
            \caption{Whiteboard Wiping: board raised.}
        \end{subfigure}

        \vspace{0.5cm}

        \begin{subfigure}{0.48\textwidth}
            \centering
            \includegraphics[width=\linewidth]{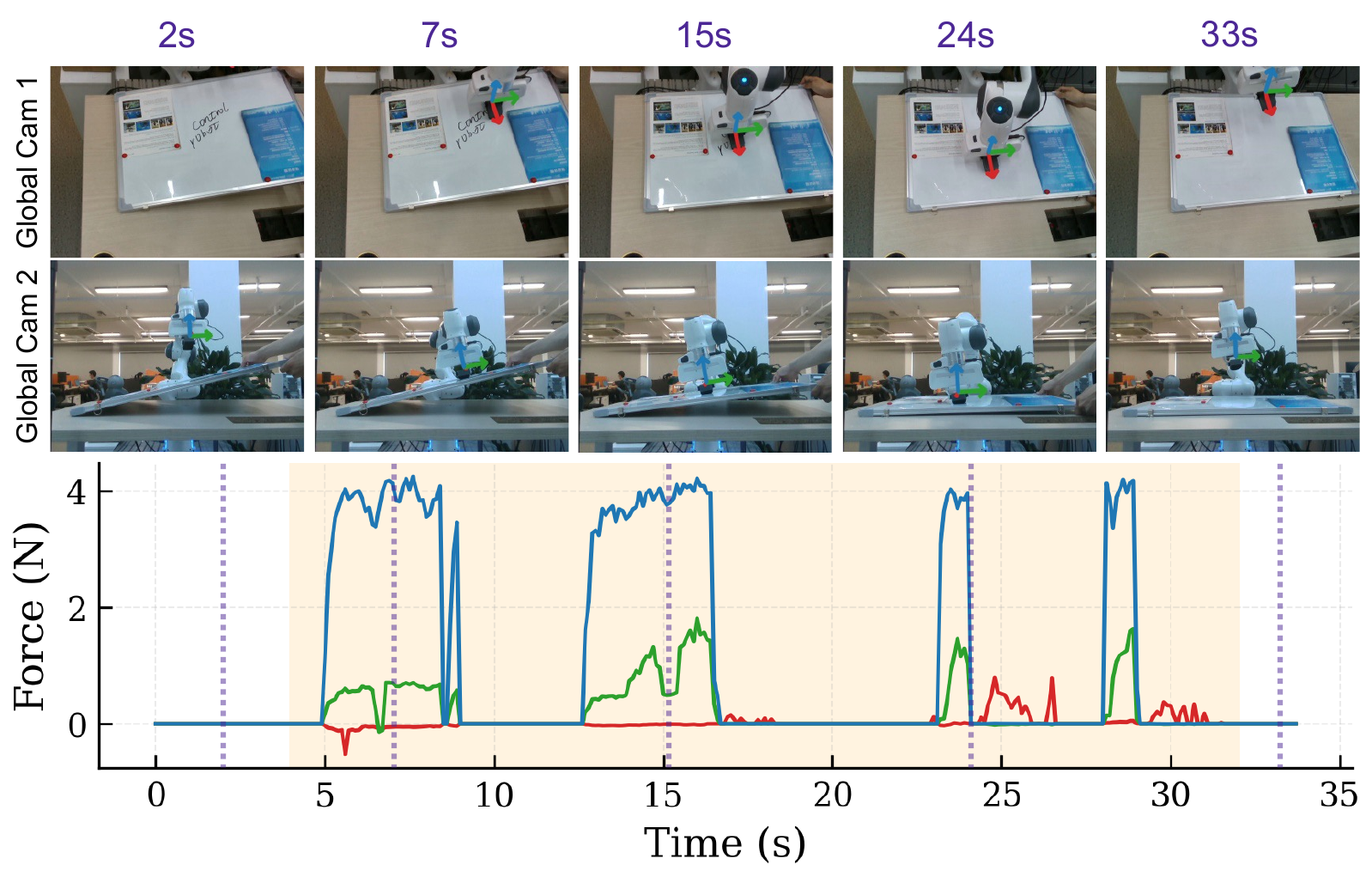}
            \vspace{-0.5cm}
            \caption{Whiteboard Wiping: board tilted.}
        \end{subfigure}
        \hfill
        \begin{subfigure}{0.48\textwidth}
            \centering
            \includegraphics[width=\linewidth]{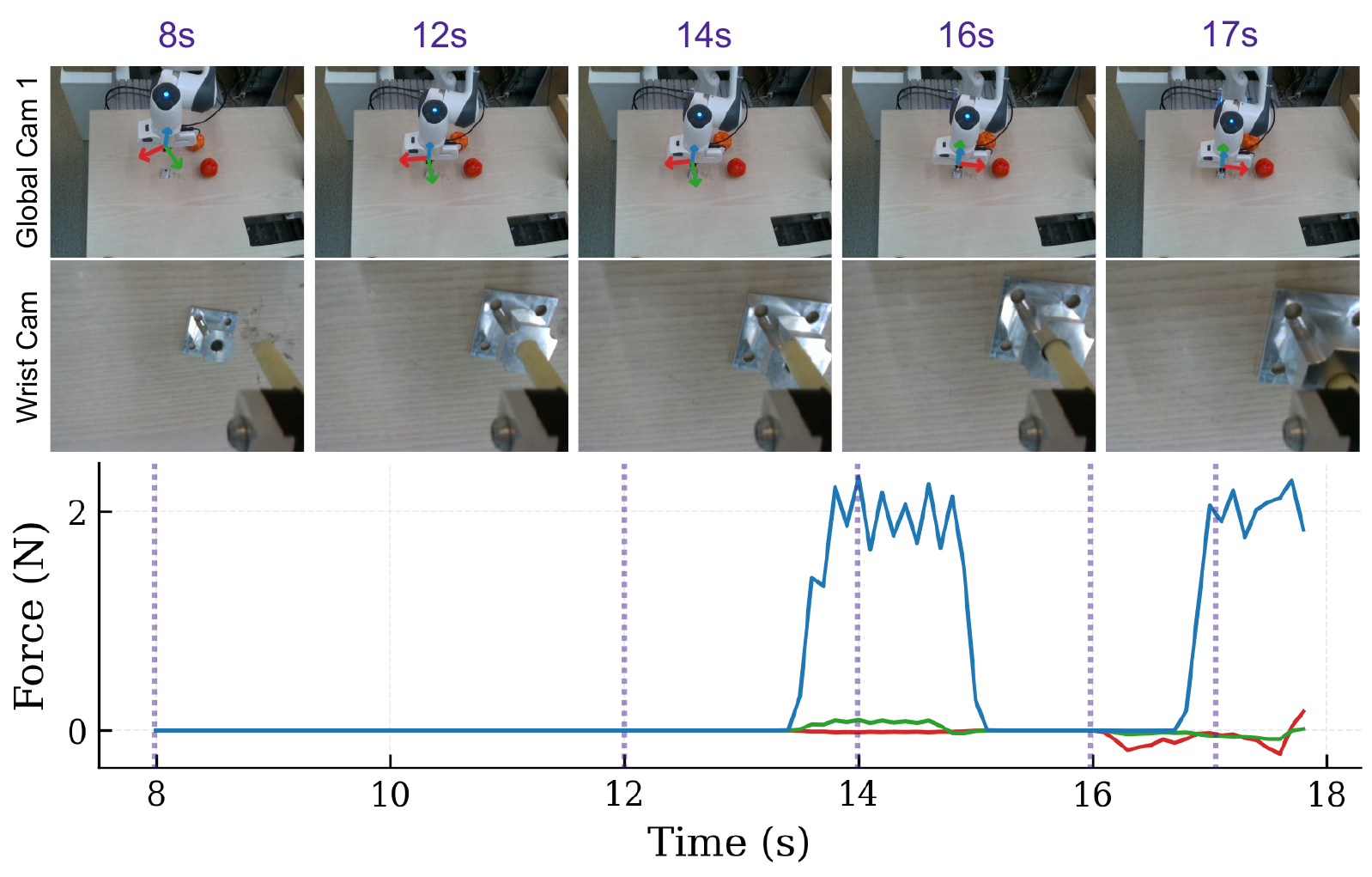}
            \vspace{-0.5cm}
            \caption{Peg-in-Hole.}
        \end{subfigure}

        \vspace{0.5cm}

        \begin{subfigure}{0.48\textwidth}
            \centering
            \includegraphics[width=\linewidth]{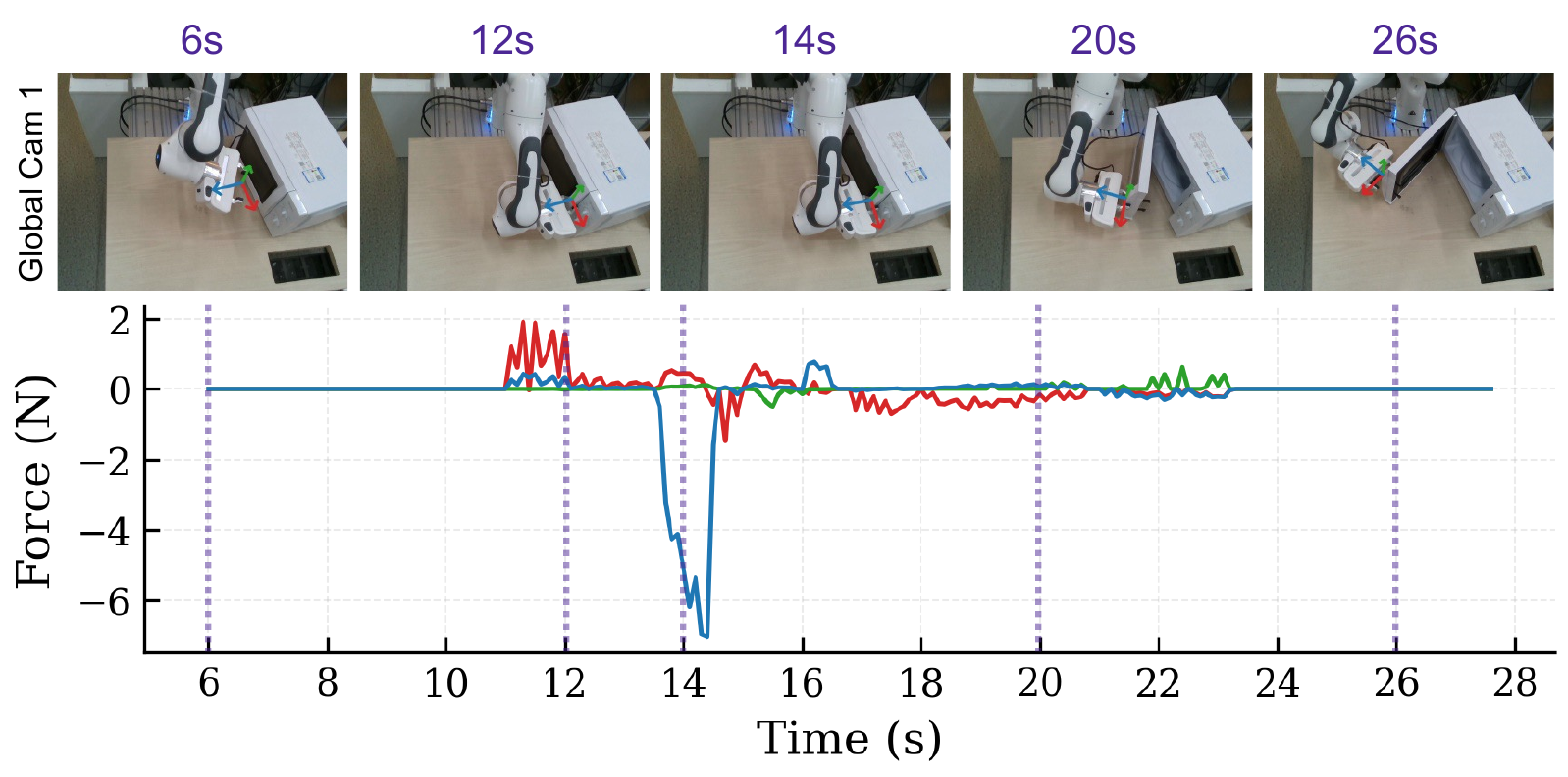}
            \vspace{-0.5cm}
            \caption{Microwave Opening: undisturbed.}
        \end{subfigure}
        \hfill
        \begin{subfigure}{0.48\textwidth}
            \centering
            \includegraphics[width=\linewidth]{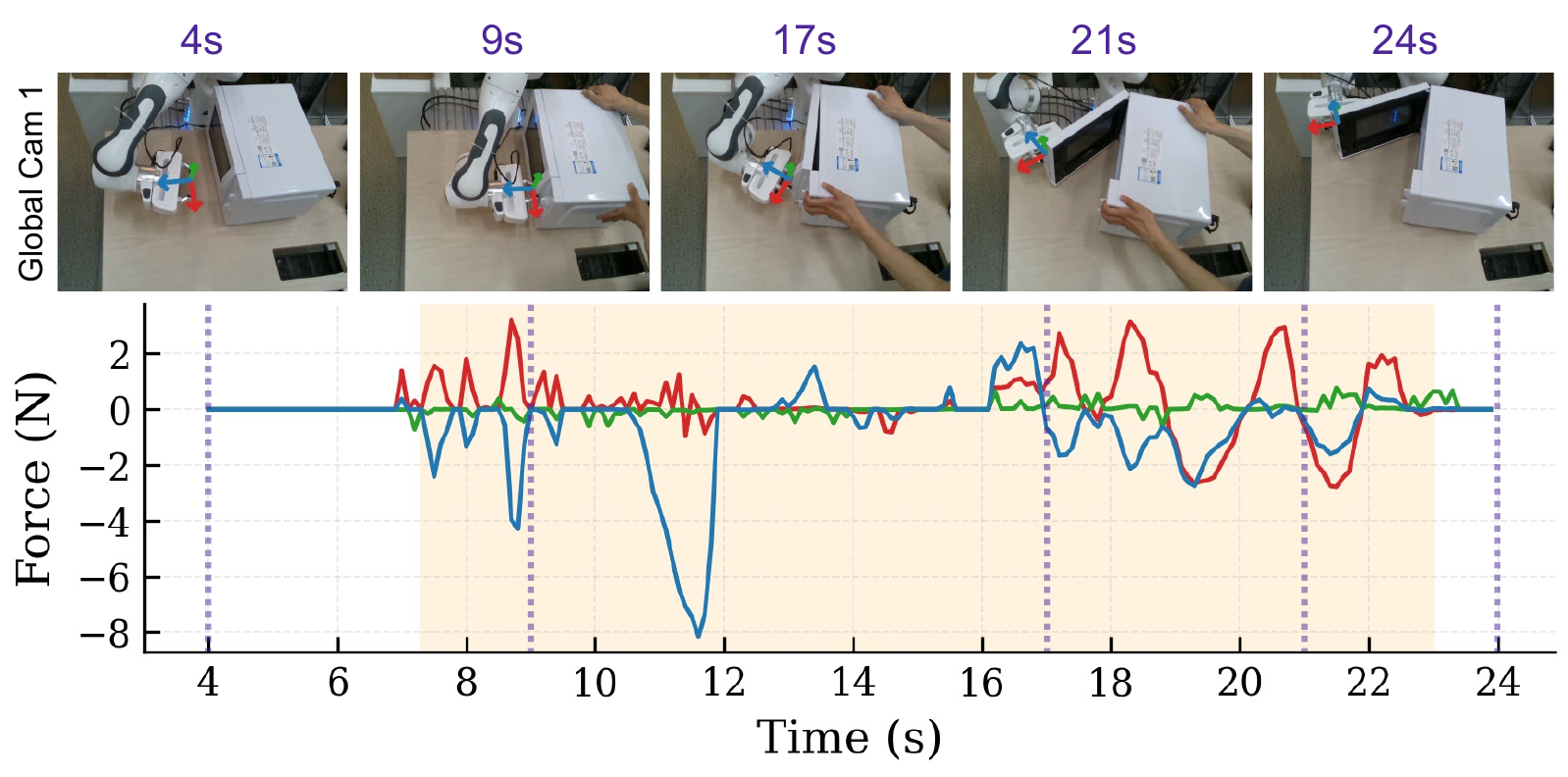}
            \vspace{-0.5cm}
            \caption{Microwave Opening: disturbed.}
        \end{subfigure}

        \vspace{0.5cm}

        \begin{subfigure}{0.48\textwidth}
            \centering
            \includegraphics[width=\linewidth]{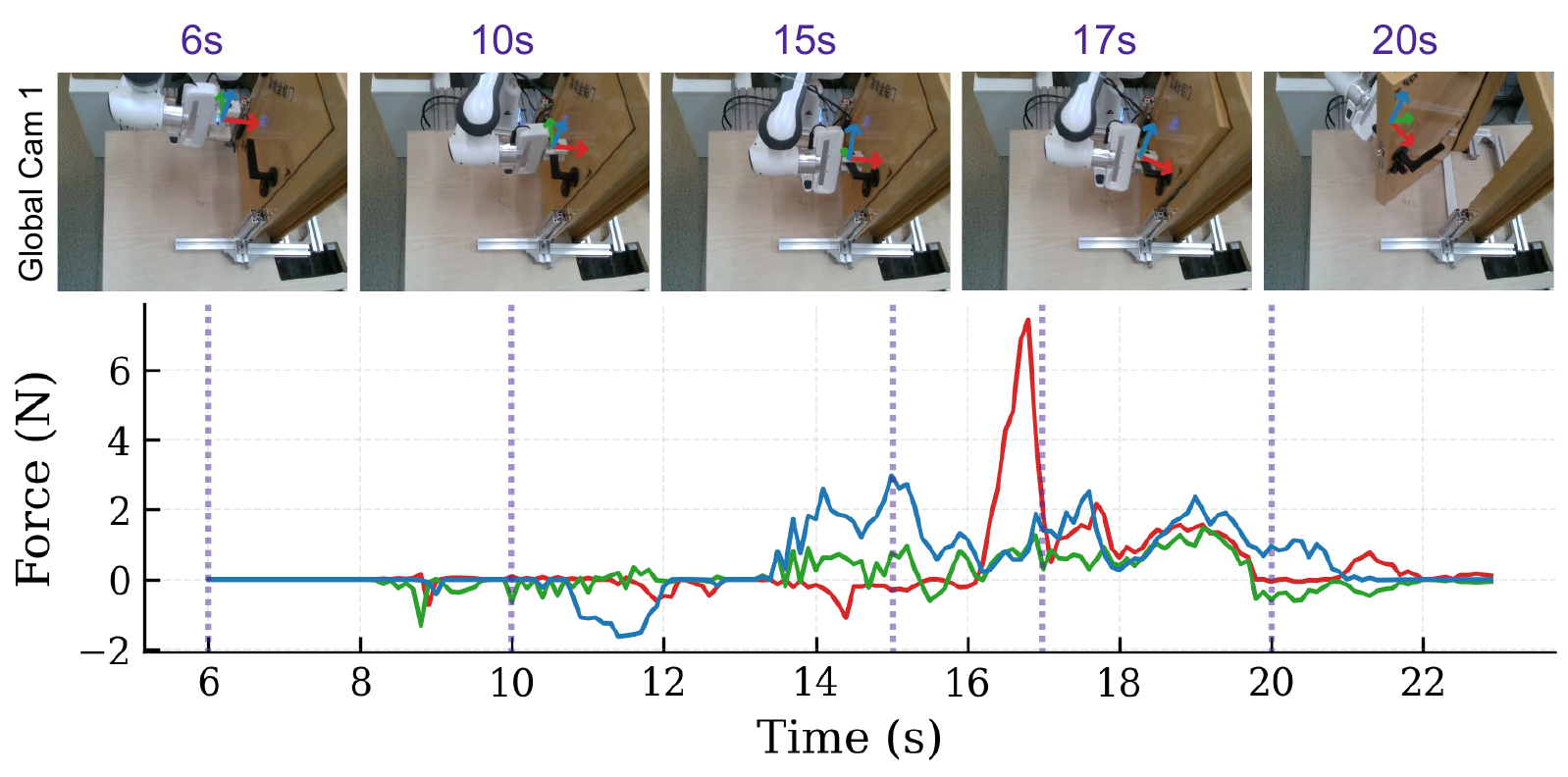}
            \vspace{-0.5cm}
            \caption{Door Opening: undisturbed.}
        \end{subfigure}
        \hfill
        \begin{subfigure}{0.48\textwidth}
            \centering
            \includegraphics[width=\linewidth]{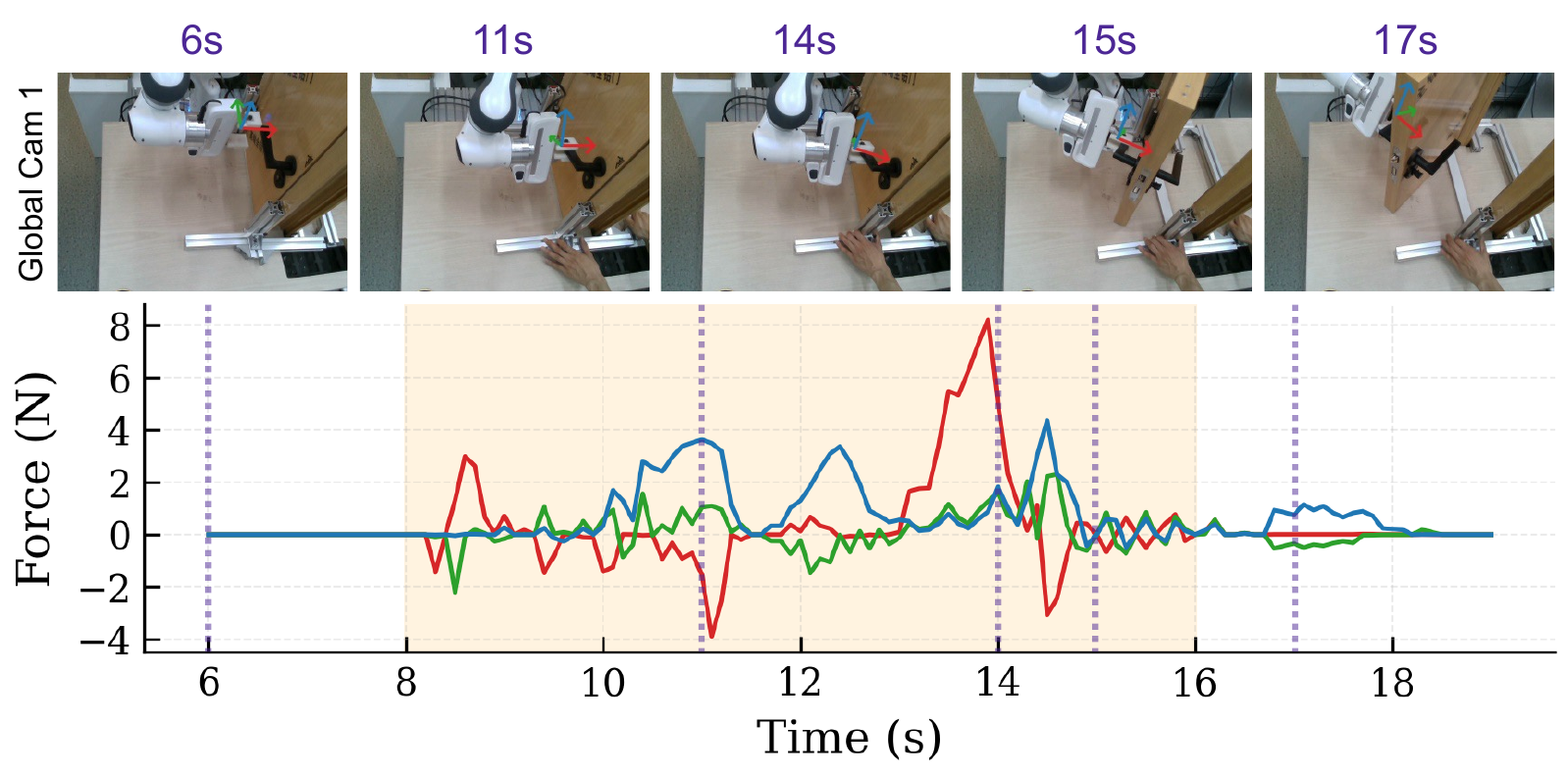}
            \vspace{-0.5cm}
            \caption{Door Opening: disturbed.}
        \end{subfigure}

        \vspace{0.5cm}

        \includegraphics[width=0.9\textwidth]{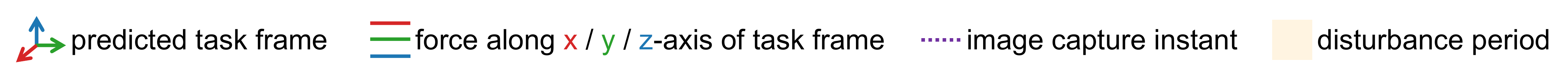}
        
        \caption{Force profiles of TDC during real-world execution. Each panel shows RGB images with the predicted task frame overlaid (top) and the corresponding three-axis force in the task frame (bottom). Purple dashed lines mark the image capture instants. Yellow shaded regions indicate disturbance periods. (a)--(c) illustrate three types of disturbances on whiteboard wiping: the board is lowered, raised, and tilted, respectively. (d) shows a full peg insertion. (e)--(f) and (g)--(h) exhibit undisturbed and disturbed trials for microwave opening and door opening, respectively.}
        \label{fig:qualitative}
    \end{figure}

    Fig.~\ref{fig:qualitative} visualizes the force profiles of TDC during real-world execution across all four tasks.

    \textbf{Whiteboard Wiping (Fig.~\ref{fig:qualitative}a--c).} In all three trials, the z-axis force (blue) rapidly converges to the 4\,N target upon contact, maintaining stable contact with the board, while the y-axis force (green) provide sufficient force to overcome wiping friction. When the board is lowered (Fig.~\ref{fig:qualitative}a) or raised (Fig.~\ref{fig:qualitative}b), the robot tracks the board's height change and maintains contact. The z-axis force deviates from the target during the disturbance but converges back rapidly once the disturbance is removed. When the board is tilted (Fig.~\ref{fig:qualitative}c), the task frame reorients to stay aligned with the new board surface, and the z-axis force remains at 4\,N throughout, maintaining stable contact across different tilt angles.

    \textbf{Peg-in-Hole (Fig.~\ref{fig:qualitative}d).} Upon contact, the z-axis force converges to the 2\,N target along the insertion axis, pressing the peg to full insertion depth, while the x-axis and y-axis forces stay small, indicating successful lateral alignment. The predicted task frame exhibits rotational jumps about the z-axis, which is expected since the hole is rotationally symmetric about the insertion axis and the task frame orientation around this axis is not uniquely defined.

    \textbf{Microwave Opening (Fig.~\ref{fig:qualitative}e--f).} In the undisturbed trial (Fig.~\ref{fig:qualitative}e), the force profile shows that sufficient force is applied along the direction (blue) that must overcome the door latch resistance, while the remaining axes stay compliant. In the disturbed trial (Fig.~\ref{fig:qualitative}f), the microwave is displaced during opening. The compliant axes absorb the disturbance, the task frame adjusts to the new microwave pose, and the robot continues to open the door successfully.

    \textbf{Door Opening (Fig.~\ref{fig:qualitative}g--h).} In the undisturbed trial (Fig.~\ref{fig:qualitative}g), force is first applied along the z-axis (blue) to turn the handle, then along the x-axis (red) to pull the door open while the z-axis force is maintained to hold the handle position. In the disturbed trial (Fig.~\ref{fig:qualitative}h), the door is displaced mid-execution. The compliant axes absorb the perturbation while the task frame realigns with the displaced door, allowing the robot to complete the task.

\end{document}